\documentclass[10pt,twocolumn,letterpaper]{article}

\usepackage[pagenumbers]{cvpr} %

\usepackage{times}
\usepackage{epsfig}
\usepackage{graphicx}
\usepackage{amsmath}
\usepackage{amssymb}
\usepackage{booktabs}
\usepackage{bm}
\usepackage{bbm}
\usepackage{setspace}
\usepackage{caption}
\usepackage{multirow}
\usepackage{extarrows}
\usepackage[normalem]{ulem}

\graphicspath{{./figures/}}

\makeatletter
\let\MYcaption\@makecaption
\makeatother

\usepackage{subcaption}

\makeatletter
\let\@makecaption\MYcaption
\makeatother

\usepackage[pagebackref,breaklinks,colorlinks]{hyperref}

\usepackage[capitalize]{cleveref}
\crefname{section}{Sec.}{Secs.}
\Crefname{section}{Section}{Sections}
\Crefname{table}{Table}{Tables}
\crefname{table}{Tab.}{Tabs.}

\newcommand{\z}{\bm{z}}
\newcommand{\zpls}{\bm{z}^{+}}
\newcommand{\zmpls}{\bm{z}_{M+}}
\newcommand{\fz}{\bm{z}^{\ast}}
\newcommand{\f}{\bm{f}}
\newcommand{\w}{\bm{w}}
\newcommand{\wpls}{\bm{w}^{+}}
\newcommand{\wmpls}{\bm{w}_{M+}}
\newcommand{\fw}{\bm{w}^{\ast}}
\newcommand{\PS}{\mathcal{P}}
\newcommand{\PNS}{\mathcal{P}_{\mathcal{N}}}
\newcommand{\PPNS}{\mathcal{P}^{+}_{\mathcal{N}}}
\newcommand{\FWS}{\mathcal{F}/\mathcal{W}^{+}}
\newcommand{\WS}{\mathcal{W}}
\newcommand{\WPS}{\mathcal{W}^{+}}
\newcommand{\ZS}{\mathcal{Z}}
\newcommand{\ZPS}{\mathcal{Z}^{+}}
\newcommand{\FZS}{\mathcal{F}/\mathcal{Z}^{+}}
\newcommand{\FSS}{\mathcal{F}/\mathcal{S}}
\newcommand{\SSp}{\mathcal{S}}
\newcommand{\FS}{\mathcal{F}}
\newcommand{\FZ}{\mathcal{F}/\mathcal{Z}}
\newcommand{\X}{\mathcal{X}}
\newcommand{\x}{\bm{x}}

\def\cvprPaperID{20} %
\def\confName{CVPR}
\def\confYear{2023}

\begin{document}

\title{Balancing Reconstruction and Editing Quality of GAN Inversion \\ for Real Image Editing with StyleGAN Prior Latent Space}

\author{Kai Katsumata\\
The University of Tokyo, Japan\\
{\tt\small katsumata@nlab.ci.i.u-tokyo.ac.jp}\\
\and
Duc Minh Vo\\
The University of Tokyo, Japan\\
{\tt\small vmduc@nlab.ci.i.u-tokyo.ac.jp}\\
\and
Bei Liu\\
Microsoft Research Asia, China\\
{\tt\small bei.liu@microsoft.com}\\
\and
Hideki Nakayama\\
The University of Tokyo, Japan\\
{\tt\small nakayama@nlab.ci.i.u-tokyo.ac.jp}\\
}

\maketitle

\begin{abstract}
The exploration of the latent space in StyleGANs and GAN inversion exemplify impressive real-world image editing, yet the trade-off between reconstruction quality and editing quality remains an open problem.
In this study, we revisit StyleGANs' hyperspherical prior $\ZS$ and $\ZPS$ and 
integrate them into seminal GAN inversion methods to improve editing quality. Besides faithful reconstruction, our extensions achieve sophisticated editing quality with the aid of the StyleGAN prior.
We project the real images into the proposed space to obtain the inverted codes, by which we then move along $\ZPS$, enabling semantic editing without sacrificing image quality.
Comprehensive experiments 
show that $\ZPS$ can replace the most commonly-used $\WS$, $\WPS$, and $\SSp$ spaces while preserving reconstruction quality, 
resulting in reduced distortion of edited images.

\end{abstract}

\section{Introduction}
\label{sec:intro}
    
The combination of GAN inversion~\cite{abdal2019image2stylegan,abdal2020image2stylegan++,Kang_2021_ICCV,feng2022near,roich2021pivotal,xia2022gan,parmar2022spatially,zhu2021barbershop,bermano2022state,zhu2020domain} and latent space editing~\cite{Shen_2020_CVPR,NEURIPS2020_ganspace,Shen_2021_CVPR} enables us to edit a wide range of image attributes such as aging, expression, and light condition, by applying editing operations~\cite{Shen_2020_CVPR,NEURIPS2020_ganspace,Shen_2021_CVPR} to inverted latent codes.
To this end, many methods
~\cite{abdal2019image2stylegan,abdal2020image2stylegan++,Kang_2021_ICCV,feng2022near} aiming to find the latent code of StyleGANs~\cite{Karras2019style,Karras2020analyzing,Karras2020training,Karras2021alias} that generates a given image have been developed. Recent efforts~\cite{tov2021designing,Kang_2021_ICCV,feng2022near,wu2021stylespace} improves
reconstruction quality of both in-domain and out-of-domain images.

One of remaining challenges of GAN inversion is the trade-off between reconstruction quality and perceptual quality of the edited images.
Popular latent spaces such as $\WS$~\cite{Karras2020analyzing}, $\WPS$~\cite{abdal2019image2stylegan}, and $\SSp$~\cite{wu2021stylespace} improve reconstruction quality, yet the low-quality edited image is unavoidable.
Recent attempts (\eg, SAM~\cite{parmar2022spatially}, PTI~\cite{roich2021pivotal}, and $\PS$~\cite{zhu2020improved}) aim to maintain the perceptual quality during semantic editing.
However, since they use transformed spaces such as $\WS$ or $\WPS$, namely unbounded space with unknown boundaries, the shape of such space is too complex to edit.
Such an unbounded embedding space cannot guarantee that the edited embedding is always present in the embedding space, resulting in distortion after editing.
Unlike typically used spaces, StyleGAN's prior space $\ZS$ is a bounded embedding space, meaning that it is rich in editing quality while poor in reconstruction quality.
To address the mentioned challenge, one solution is to flexibly use the rich embedding space and the original latent space, which is our main target. 

To begin with, we revisit the original latent space $\ZS$, which can be easily editable. Since the latent code $\z \in \ZS$ is sampled from the hypersphere, 
the latent code can move on $\ZS$ with closed operations. 
To maintain the reconstruction quality while leveraging the robust nature of $\ZS$, we first extend $\ZS$ to $\ZPS$ and then combine $\ZPS$ with a feature space $\FS$ in the output of an intermediate layer in the StyleGAN generator, proposing an alternative space namely $\FZS$.
Our proposed $\FZS$ space achieves excellent reconstruction performance due to the use of the feature space $\FS$ and increases editing quality with the aid of the hyperspherical prior of the original latent space $\ZPS$, simultaneously. Here, editing quality denotes the perceptual quality of images after performing editing operations in the latent space.

Qualitative and quantitative evaluations show that our method maintains image quality after performing editing operations without sacrificing reconstruction quality.
We futher demonstrate that the our method can be applied to many cutting-edge GAN inversion methods.

 \begin{figure}[tb]
   \centering
   \begin{minipage}{0.38\linewidth}
      \includegraphics[height=17em]{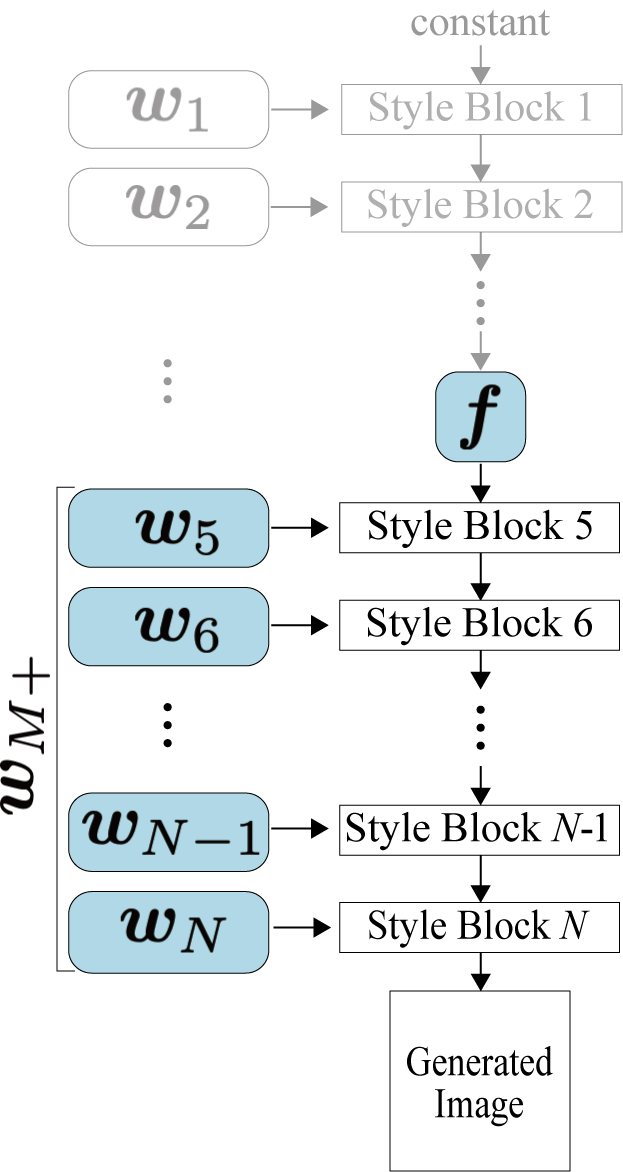}
      \subcaption{$\mathcal{F}/\mathcal{W}^{+}$ space~\cite{Kang_2021_ICCV} \label{fig:fwpspace}}
   \end{minipage}%
   \hfill
   \begin{minipage}{0.62\linewidth}
      \includegraphics[height=17em]{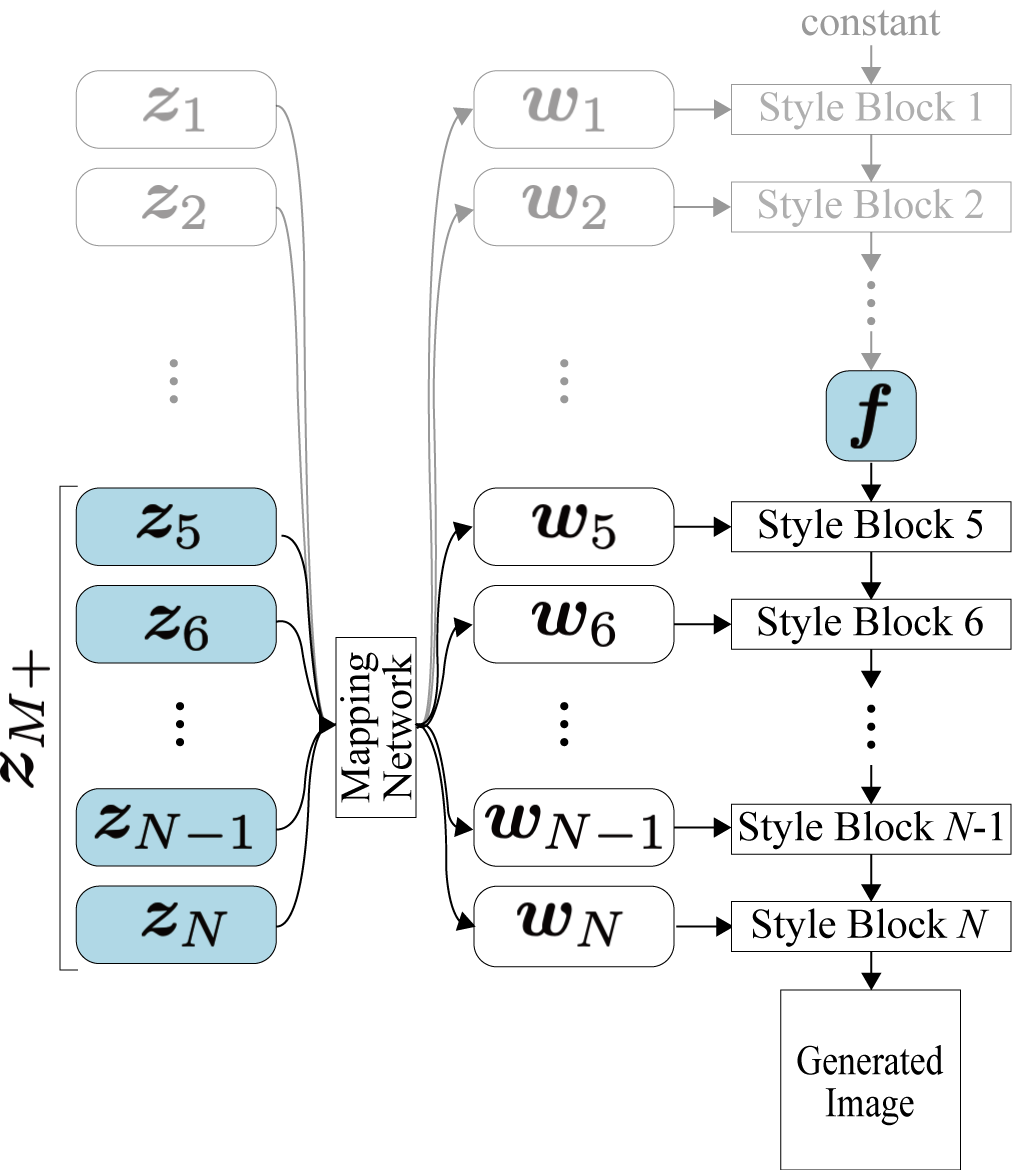}
      \subcaption{$\mathcal{F}/\mathcal{Z}^{+}$ space (ours)\label{fig:fzpspace}}
   \end{minipage}%
\vspace{-0.4em}
   \caption{Latent spaces of StyleGANs. The space $\FWS$
     leads to the faithful reconstruction. Using $\ZPS$ instead of $\WPS$, $\FZS$ does not sacrifice reconstruction quality with the aid of $\FS$.
    The base code $\f$ is an intermediate output of the StyleGAN generator with spatial dimensions, and 
    the detail code $\wmpls$ or $\zmpls$ is a subset of
    $\wpls$ or $\zpls$ and the inputs of the upper stages of the generator. The optimizing codes are highlighted in blue. \vspace{-1em}}
\end{figure}

\section{Approach}\label{sec:method}

In this section, we first review various latent spaces for GAN inversion
and their pros and cons.
Then, we introduce the integration of $\ZPS$ into cutting-edge GAN inversion methods for improving editing quality while maintaining reconstruction quality.

\subsection{Analysis of StyleGAN Spaces}\label{sec:fzspace}

\begin{figure}
  \centering
    \bgroup 
    \def\arraystretch{0.2} 
    \setlength\tabcolsep{0.2pt}
    \begin{tabular}{cccccc}
\raisebox{1.6em}{target} & 
\includegraphics[width=0.168\columnwidth]{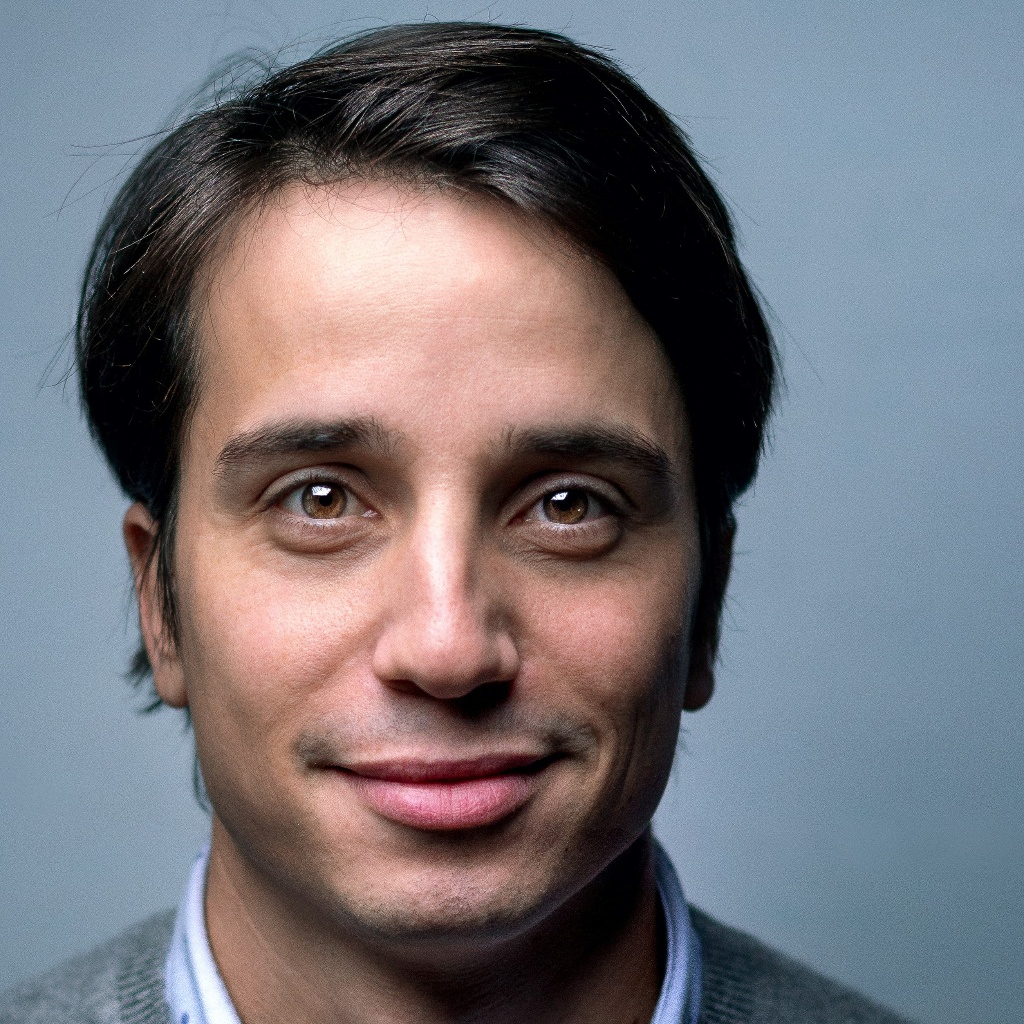} &
\includegraphics[width=0.168\columnwidth]{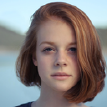} &
\includegraphics[width=0.168\columnwidth]{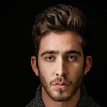} &
\includegraphics[width=0.168\columnwidth]{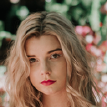} &
\includegraphics[width=0.168\columnwidth]{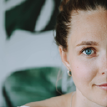} \\
\raisebox{1.6em}{\begin{tabular}{c}$\mathcal{F}/\mathcal{W}^{+}$ \\ \!$\!\!(\PNS\!)\!$\cite{Kang_2021_ICCV}\!\end{tabular}} & 
\includegraphics[width=0.168\columnwidth]{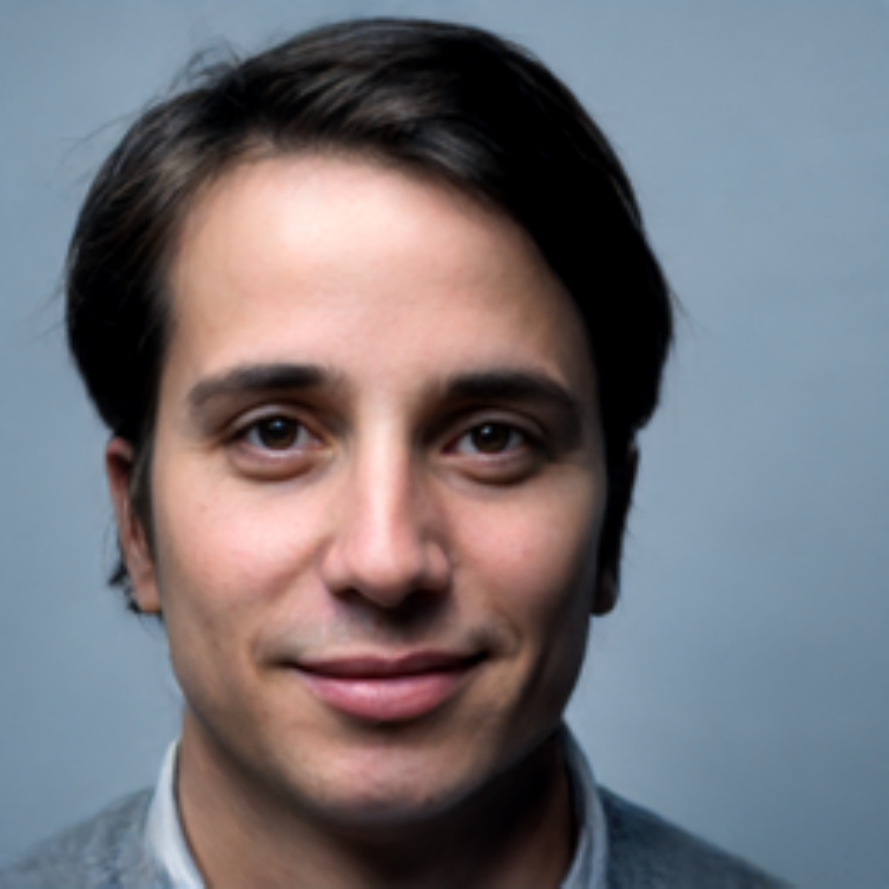} &
\includegraphics[width=0.168\columnwidth]{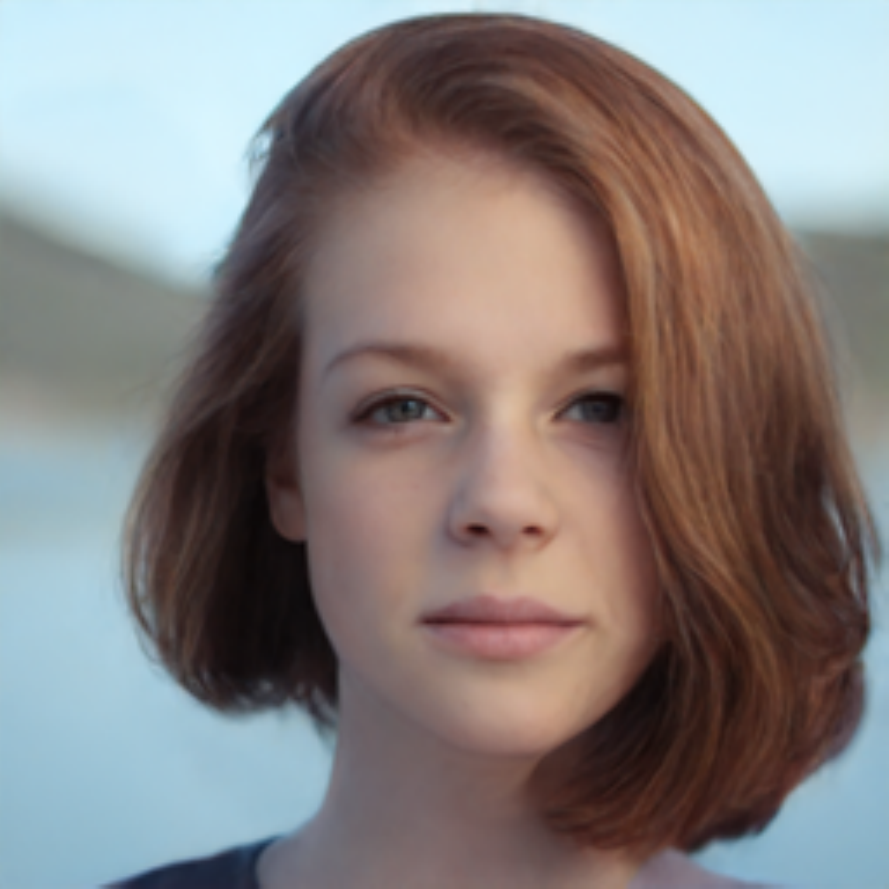} &
\includegraphics[width=0.168\columnwidth]{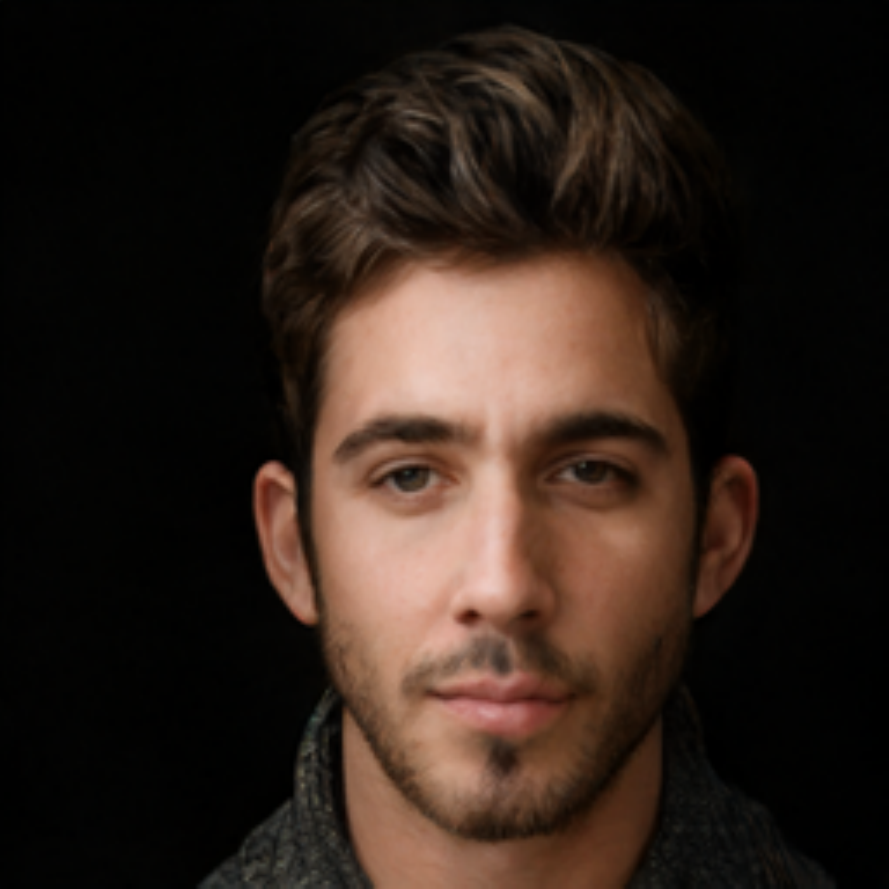} &
\includegraphics[width=0.168\columnwidth]{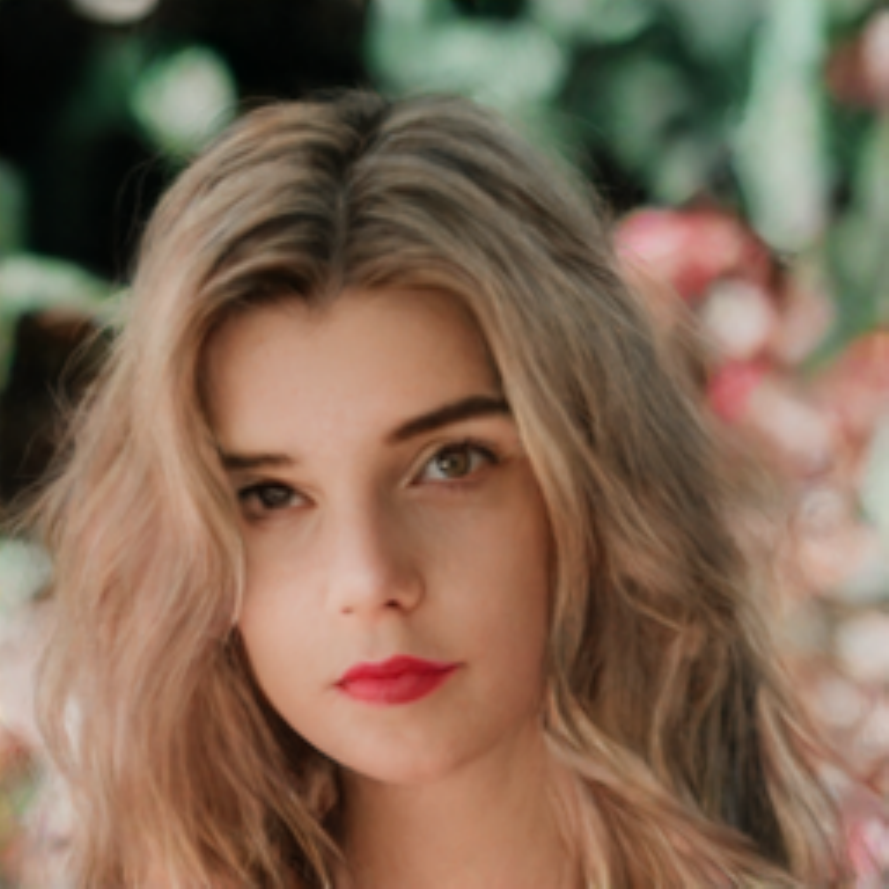} &
\includegraphics[width=0.168\columnwidth]{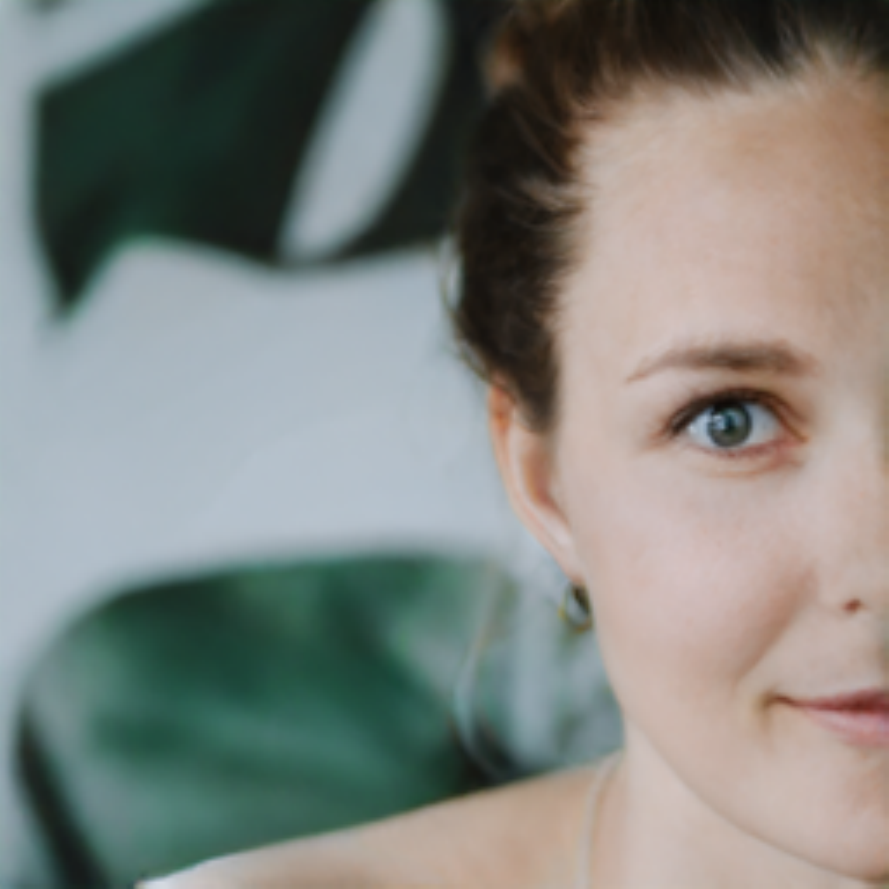} \\
\footnotesize LPIPS loss & \footnotesize 0.05944 & \footnotesize 0.06785 & \footnotesize 0.06728 & \footnotesize 0.07973 & \footnotesize 0.09912 \\
\raisebox{1.2em}{\!$\mathcal{F}\!/\!\mathcal{S}$\!~\cite{yao2022style}\!} & 
\includegraphics[width=0.168\columnwidth]{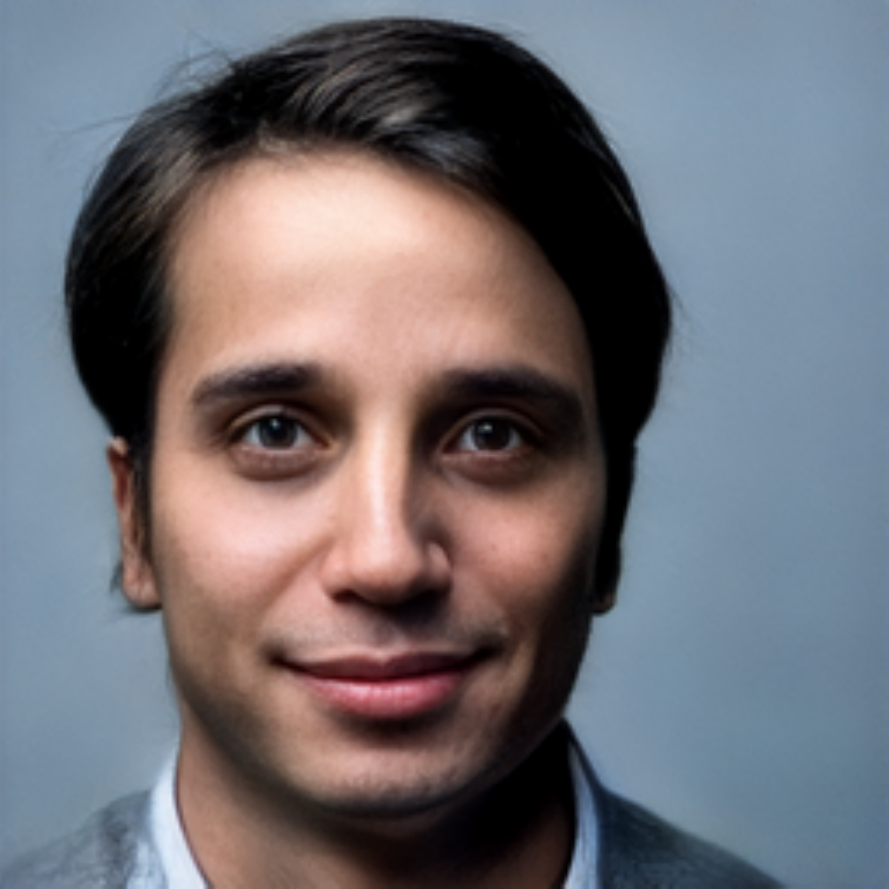} &
\includegraphics[width=0.168\columnwidth]{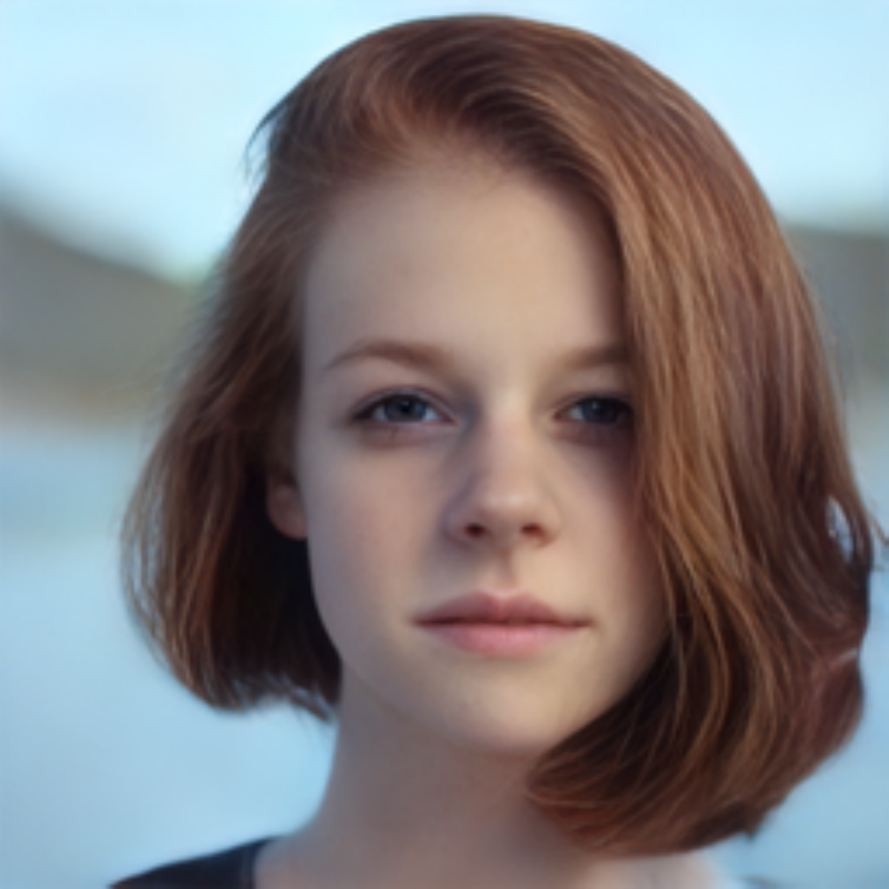} &
\includegraphics[width=0.168\columnwidth]{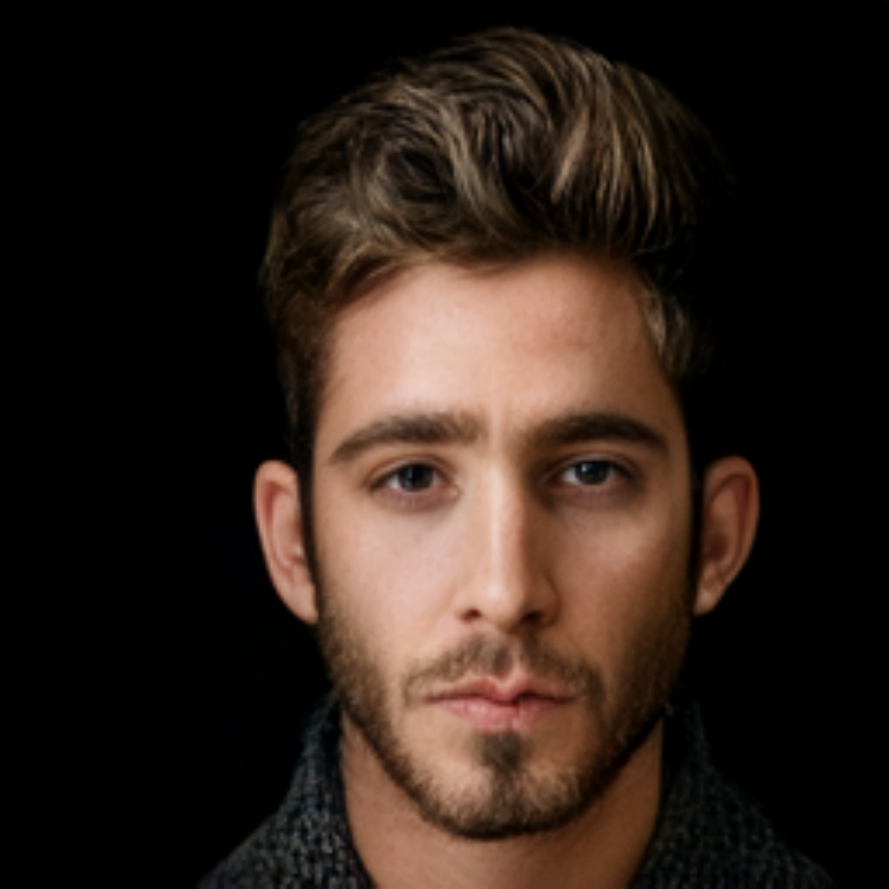} &
\includegraphics[width=0.168\columnwidth]{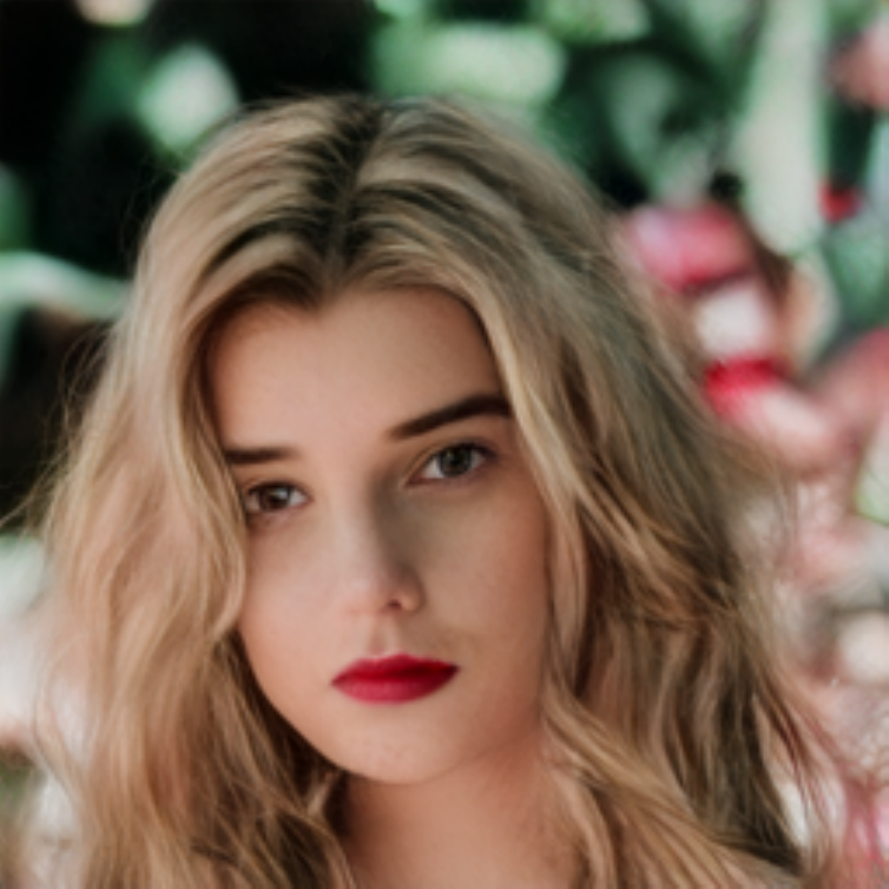} &
\includegraphics[width=0.168\columnwidth]{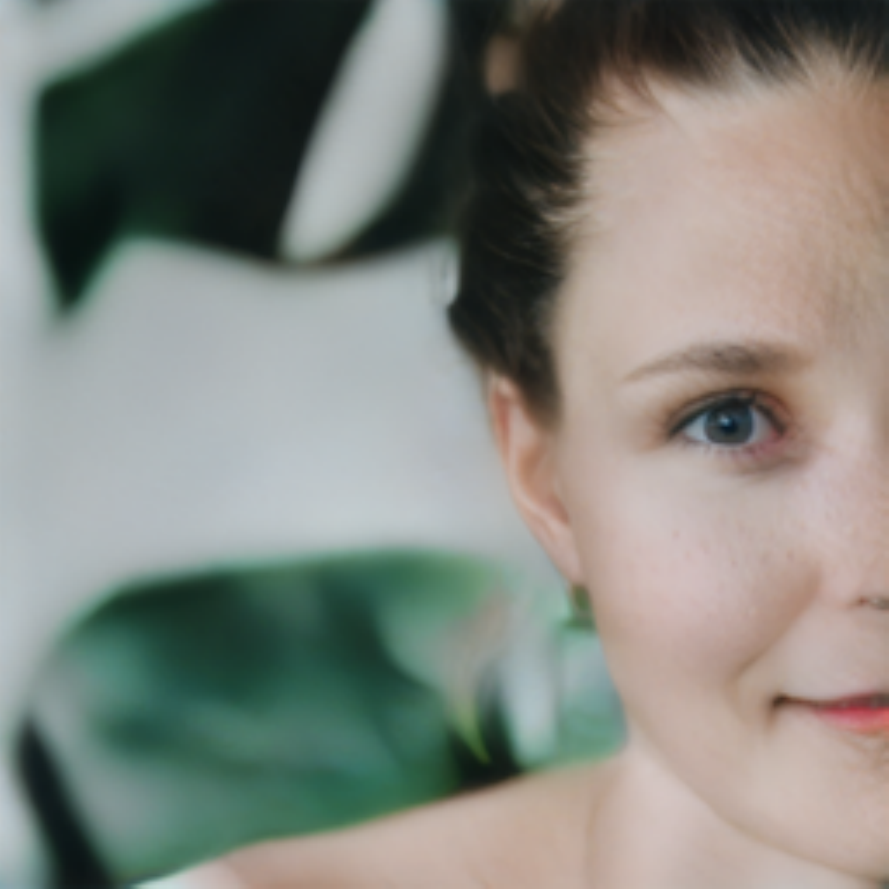} \\
\footnotesize LPIPS loss & \footnotesize 0.10868 & \footnotesize 0.07482 & \footnotesize 0.43191 & \footnotesize 0.12273 & \footnotesize $\textbf{0.07196}$ \\
\raisebox{1.2em}{\shortstack{$\mathcal{F}/\mathcal{Z}$ \\ (Ours)}} & 
\includegraphics[width=0.168\columnwidth]{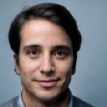} &
\includegraphics[width=0.168\columnwidth]{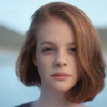} &
\includegraphics[width=0.168\columnwidth]{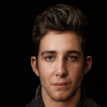} &
\includegraphics[width=0.168\columnwidth]{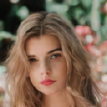} &
\includegraphics[width=0.168\columnwidth]{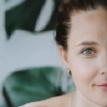} \\
\footnotesize LPIPS loss & \footnotesize 0.06573 & \footnotesize 0.07567 & \footnotesize 0.08897 & \footnotesize 0.08939 & \footnotesize 0.11237 \\
\raisebox{1.2em}{\shortstack{$\mathcal{F}/\mathcal{Z}^{+}$ \\ (Ours)}} & 
\includegraphics[width=0.168\columnwidth]{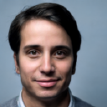} &
\includegraphics[width=0.168\columnwidth]{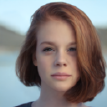} &
\includegraphics[width=0.168\columnwidth]{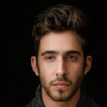} &
\includegraphics[width=0.168\columnwidth]{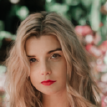} &
\includegraphics[width=0.168\columnwidth]{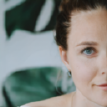} \\
\footnotesize LPIPS loss & \footnotesize $\textbf{0.05315}$ & \footnotesize $\textbf{0.06540}$ & \footnotesize $\textbf{0.06563}$ & \footnotesize $\textbf{0.07737}$ & \footnotesize 0.09158 \\
    \end{tabular}\egroup
\vspace{-0.4em}
    \caption{Comparison of inverted images with different latent spaces.
      $\FZS$ achieves high-quality reconstructions on par $\FWS$ and $\FSS$ qualitatively and quantitatively.\vspace{-2.0em}}\label{fig:recon}
\end{figure}

\vspace{-0.4em}
\begin{table}
\centering
\caption{Quantitative comparison of reconstruction quality. Our $\FZS$ yields performance comparable to $\FWS(\PNS)$.\vspace{-0.4em}}\label{tb:recon}
\bgroup
\setstretch{0.6}
\resizebox{0.96\linewidth}{!}{
\begin{tabular}{ccccc}\toprule
\!Space & $\mathcal{Z}$ & $\mathcal{Z}^{+}$ & $\mathcal{W}^{+} (P_N)$ & $\mathcal{W}^{+}$  \\\midrule
\!MSE & 0.18149 & 0.12117  & 0.11965 & 0.04872  \\
\!SSIM  & 0.61155 & 0.68930  & 0.68190 & 0.76101  \\ \midrule
\!Space & \!$\mathcal{F}\!/\!\mathcal{Z}$\! & \!$\mathcal{F}\!/\!\mathcal{Z}^{+}$\! (Ours)\!& \!$\mathcal{F}\!/\!\mathcal{W}^{+} \!(P_N)$\cite{Kang_2021_ICCV}\!\!  & \!IDInvert\cite{zhu2020domain}\!\! \\ \midrule
\!MSE & 0.02679 & $\textbf{0.01742}$ & $\textbf{0.01743}$ & 0.02155\\ 
\!SSIM & 0.78965 & $\textbf{0.81477}$   & $\textbf{0.81479}$ & 0.64993 \\\bottomrule
\end{tabular}}
\egroup
\end{table}
\vspace{-0.4em}

 \begin{figure*}[tb]
  \centering
    \bgroup 
    \def\arraystretch{0.2} 
    \setlength\tabcolsep{0.2pt}
    \begin{tabular}{cccccc|cccccc|c}
\small target & \small inversion & \multicolumn{2}{|c}{\small direction 1} & \multicolumn{2}{c|}{\small direction 2} & \small target & \small inversion & \multicolumn{2}{c}{\small direction 1} & \multicolumn{2}{c|}{\small direction 2}\\
\includegraphics[width=0.07\linewidth]{target_images/sample2} & 
\includegraphics[width=0.07\linewidth]{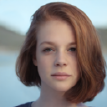} &
\includegraphics[width=0.07\linewidth]{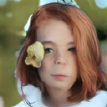} &
\includegraphics[width=0.07\linewidth]{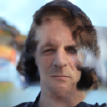} &
\includegraphics[width=0.07\linewidth]{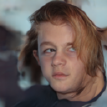} &
\includegraphics[width=0.07\linewidth]{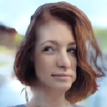} &
\includegraphics[width=0.07\linewidth]{target_images/sample4} &
\includegraphics[width=0.07\linewidth]{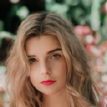} &
\includegraphics[width=0.07\linewidth]{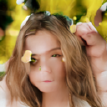} &
\includegraphics[width=0.07\linewidth]{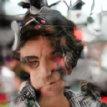} &
\includegraphics[width=0.07\linewidth]{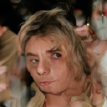} &
\includegraphics[width=0.07\linewidth]{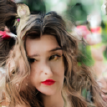} & \raisebox{1.4em}{\shortstack{\small $\mathcal{F}/\mathcal{W}^{+}$ \\$(P_N)\!$\cite{Kang_2021_ICCV}}} \\ &
\includegraphics[width=0.07\linewidth]{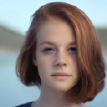} &
\includegraphics[width=0.07\linewidth]{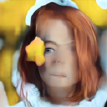} &
\includegraphics[width=0.07\linewidth]{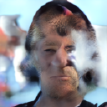} &
\includegraphics[width=0.07\linewidth]{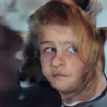} &
\includegraphics[width=0.07\linewidth]{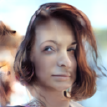} &  &
\includegraphics[width=0.07\linewidth]{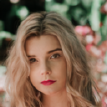} &
\includegraphics[width=0.07\linewidth]{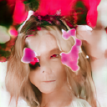} &
\includegraphics[width=0.07\linewidth]{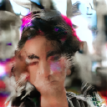} &
\includegraphics[width=0.07\linewidth]{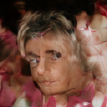} &
\includegraphics[width=0.07\linewidth]{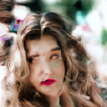} & \raisebox{1.4em}{\small $\mathcal{F}/\mathcal{W}^{+}$}\\ &
\includegraphics[width=0.07\linewidth]{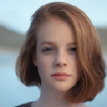} &
\includegraphics[width=0.07\linewidth]{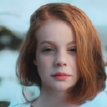} &
\includegraphics[width=0.07\linewidth]{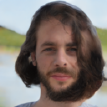} &
\includegraphics[width=0.07\linewidth]{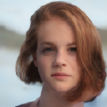} &
\includegraphics[width=0.07\linewidth]{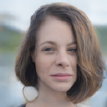} & &
\includegraphics[width=0.07\linewidth]{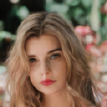} &
\includegraphics[width=0.07\linewidth]{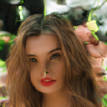} &
\includegraphics[width=0.07\linewidth]{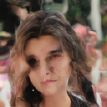} &
\includegraphics[width=0.07\linewidth]{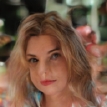} &
\includegraphics[width=0.07\linewidth]{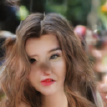} & \raisebox{1.4em}{\shortstack{\small $\mathcal{F}/\mathcal{Z}$ \\ (Ours)}} \\ &
\includegraphics[width=0.07\linewidth]{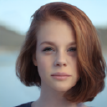} &
\includegraphics[width=0.07\linewidth]{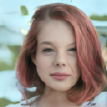} &
\includegraphics[width=0.07\linewidth]{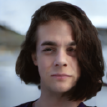} &
\includegraphics[width=0.07\linewidth]{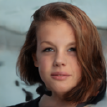} &
\includegraphics[width=0.07\linewidth]{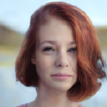} &&
\includegraphics[width=0.07\linewidth]{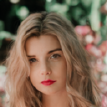} &
\includegraphics[width=0.07\linewidth]{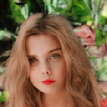} &
\includegraphics[width=0.07\linewidth]{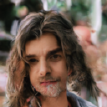} &
\includegraphics[width=0.07\linewidth]{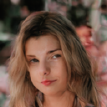} &
\includegraphics[width=0.07\linewidth]{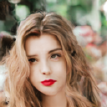} & \raisebox{1.4em}{\shortstack{\small $\mathcal{F}/\mathcal{Z}^{+}$ \\ (Ours)}}\vspace{-1.0em}
    \end{tabular}\egroup
    \caption{Editing comparison with GANSpace directions. Although the spaces with $\WPS$ fail to preserve the structure of generated faces, our spaces properly preserve them.\vspace{-2.2em}}\label{fig:editing}
\end{figure*}

\noindent \textbf{$\ZS$ and $\ZPS$ Space}.
The generator $G: \ZS \to \X$ learns to map a simple distribution, called latent space $\ZS$, to the image space, 
where $\x \in \X$ is an image, and $\z \in \ZS$ is uniformly sampled from a hypersphere.
The primitive latent code of the StyleGAN family is with $512$ dimensions.

AgileGAN~\cite{song2021agilegan} and StyleAlign~\cite{wu2021stylealign} employ the extended space $\ZPS$, which provides a different latent code from $\ZS$ for each layer.
AgileGAN~\cite{song2021agilegan} and StyleAlign~\cite{wu2021stylealign} note that $\ZS$ and $\ZPS$
have high editing quality and low reconstruction quality, and they are not suit for GAN inversion.

\noindent \textbf{$\WS$, $\WPS$, and $\SSp$ Space}. 
StyleGANs also use the intermediate latent space $\WS$ where each $\w \in \WS$ is transformed from $\z \in \ZS$ by using a mapping network. 
Thereafter, \cite{abdal2019image2stylegan,abdal2020image2stylegan++} introduced $\WPS$ space, achieving lower reconstruction loss by allowing to control of details of images.
Each element $\wpls$ in $\WPS$ is defined as $\wpls\! = \{\w_i\}_{i=1}^{N}$, where $\w_i\in\WS$, and $N$ is the number of layers in generator that takes $\w$
as input.
$\SSp$ space~\cite{wu2021stylespace} is spanned by style parameters, which is transformed from $\w \in \WS$ using different learned affine transformations for each layer of the generator.

Although the spaces derive faithful reconstruction quality, distortions and artifacts may appear in edited images~\cite{wulff2020improving,zhu2020improved,song2021agilegan}.
This is because the embeddings with these spaces for the images may not correspond appropriately to $\ZPS$, the StyleGAN prior, 
and the space cannot guarantee that the edited latent code reaches the original space.

\noindent $\PPNS$ \textbf{Space}. 
Zhu \etal\cite{zhu2020improved} introduced a normalized space $\PNS$ by whitening the logit output of the final linear layer of the mapping network.
Since the normalized space $\PNS$ can be approximated to Gaussian distribution,
penalizing the distance between the latent code and the mean of $\PNS$ locates the latent code to the high-density region.
It also can be extended to $\PPNS$ by using $\WPS$.
In the normalized spaces, editing operations are performed on $\WS$ or $\WPS$ space.

\noindent $\FWS$ \textbf{and} $\FSS$ \textbf{Space}. Kang \etal\cite{Kang_2021_ICCV} proposed $\FWS$ space
(\cref{fig:fwpspace}), consisting of the $\FS$ and $\WPS$ spaces, and the space also investigated in SAM~\cite{parmar2022spatially} and Barbershop~\cite{zhu2021barbershop}. 
The coarse-scale feature map $\f \in \FS$ is an intermediate output of a generator before taking detail codes $\wmpls = \{\w_i\}_{i=M}^{N}$.
An element $\fw = (\f, \wmpls)$ of $\FWS$ consists of a base code $\f$ and a detail code $\wmpls$. 
The information of a noise input and bottom latent codes $\{\w_i\}_{i=1}^{M-1}$ is contained in $\f$, and $\f$ controls the geometric information.
Kang \etal\cite{Kang_2021_ICCV} uses $\PPNS$. 
$\FSS$ space~\cite{yao2022style} combines $\FS$ and $\SSp$~\cite{wu2021stylespace}. 
Since they use unbounded spaces for latent editing, they have the issue of editing quality like $\WPS$.

\begin{figure}[tb]
  \centering
    \bgroup 
    \def\arraystretch{0.2} 
    \setlength\tabcolsep{0.2pt}
    \begin{tabular}{cccccccc}
\end{tabular}\\
\begin{minipage}[b]{0.18\linewidth}
\begin{tabular}{cc}
&\footnotesize $\FWS$\!~\!\cite{Kang_2021_ICCV} \\
\raisebox{1.6em}{\rotatebox[origin=c]{90}{\footnotesize inversion}} &
\includegraphics[width=0.889\linewidth]{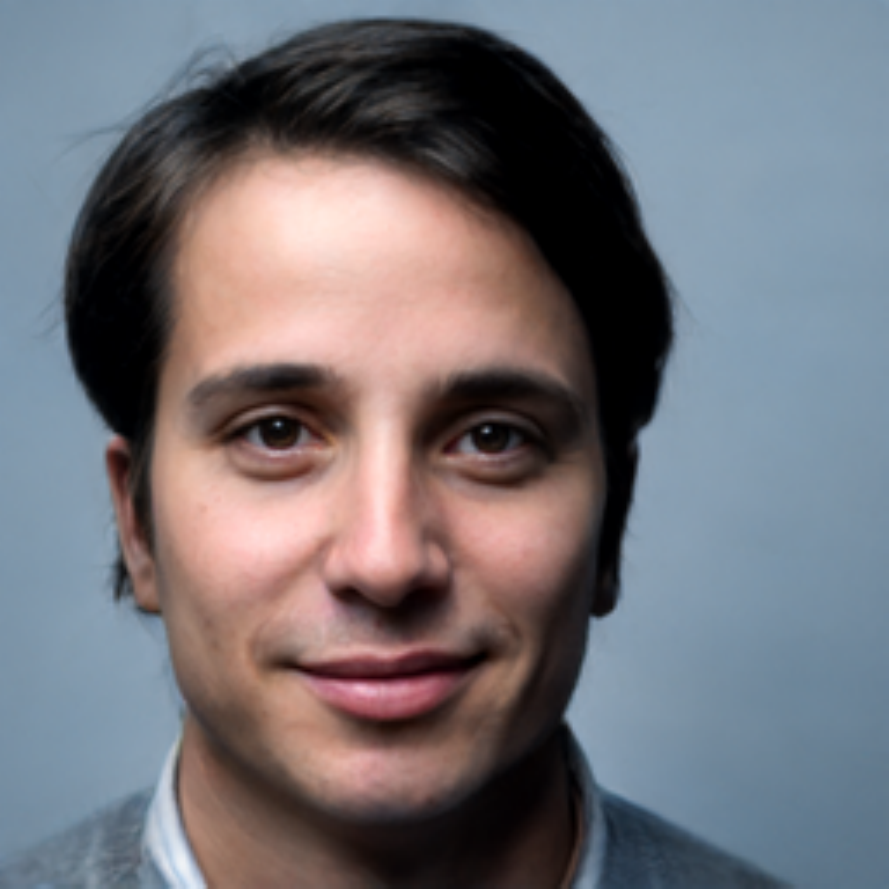} \\
\raisebox{1.7em}{\rotatebox[origin=c]{90}{\footnotesize makeup}} &
\includegraphics[width=0.889\linewidth]{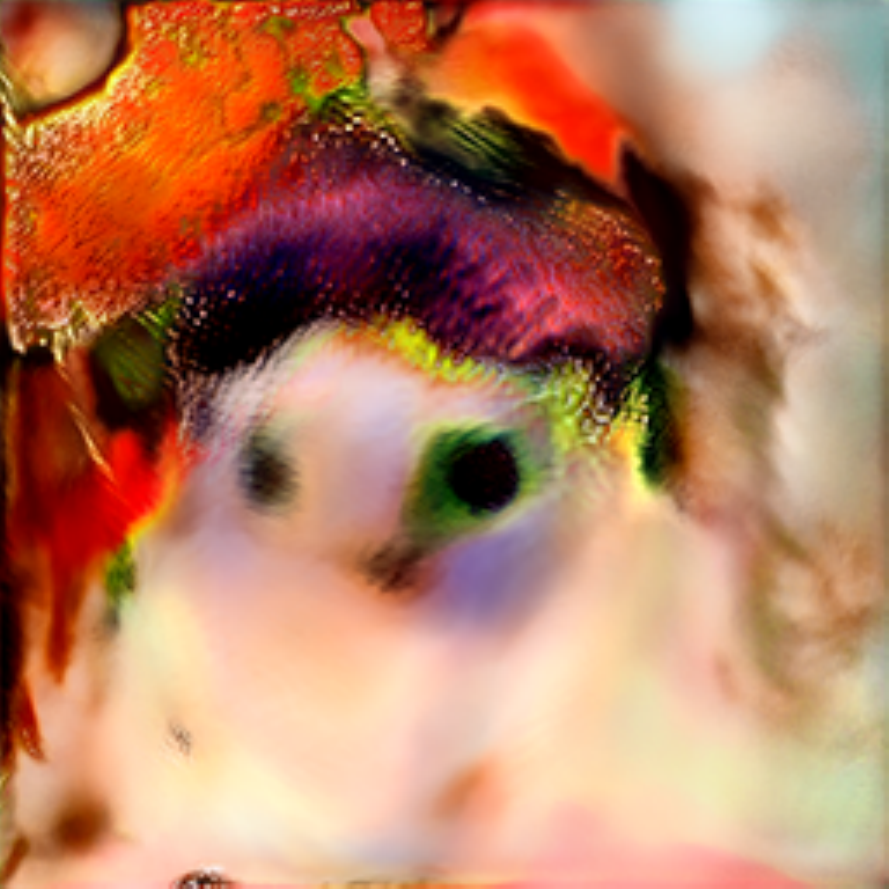} \\
\raisebox{1.8em}{\rotatebox[origin=c]{90}{\footnotesize smiling}} &
\includegraphics[width=0.889\linewidth]{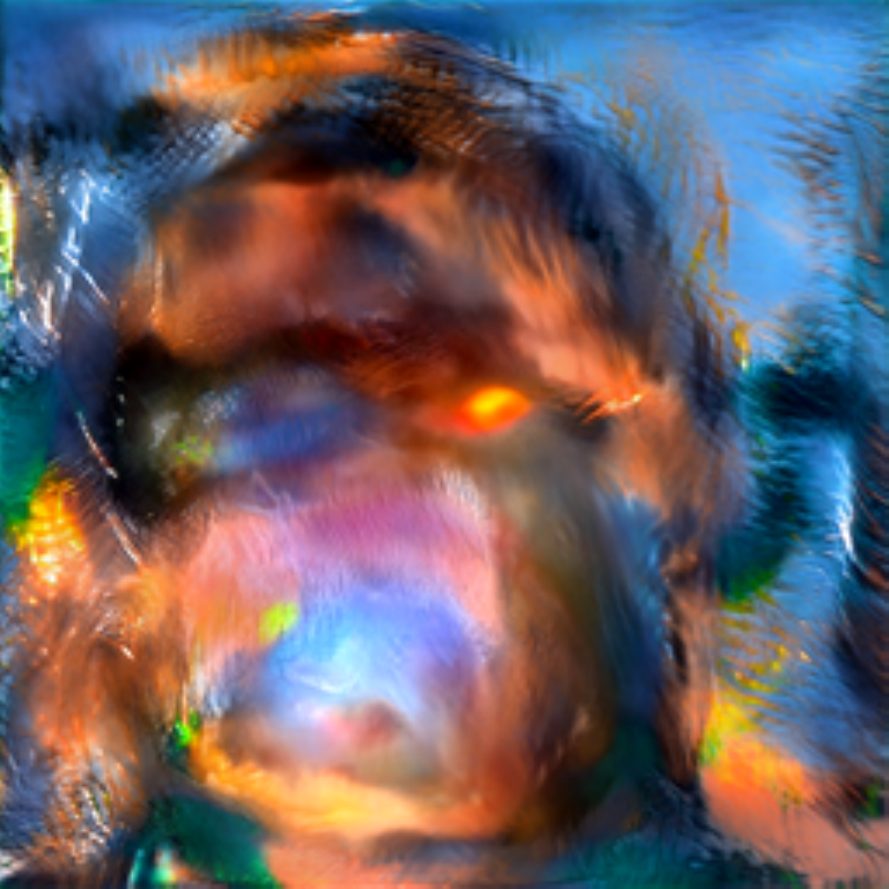} \\
\end{tabular}
\end{minipage}%
\begin{minipage}[b]{0.02\linewidth}
\end{minipage}%
\begin{minipage}[b]{0.16\linewidth}
\begin{tabular}{c}
\footnotesize $\mathcal{F}\!/\!\mathcal{S}$~\cite{yao2022style} \\
\includegraphics[width=\linewidth]{interfacegan/sp_sample1_1.pdf} \\
\includegraphics[width=\linewidth]{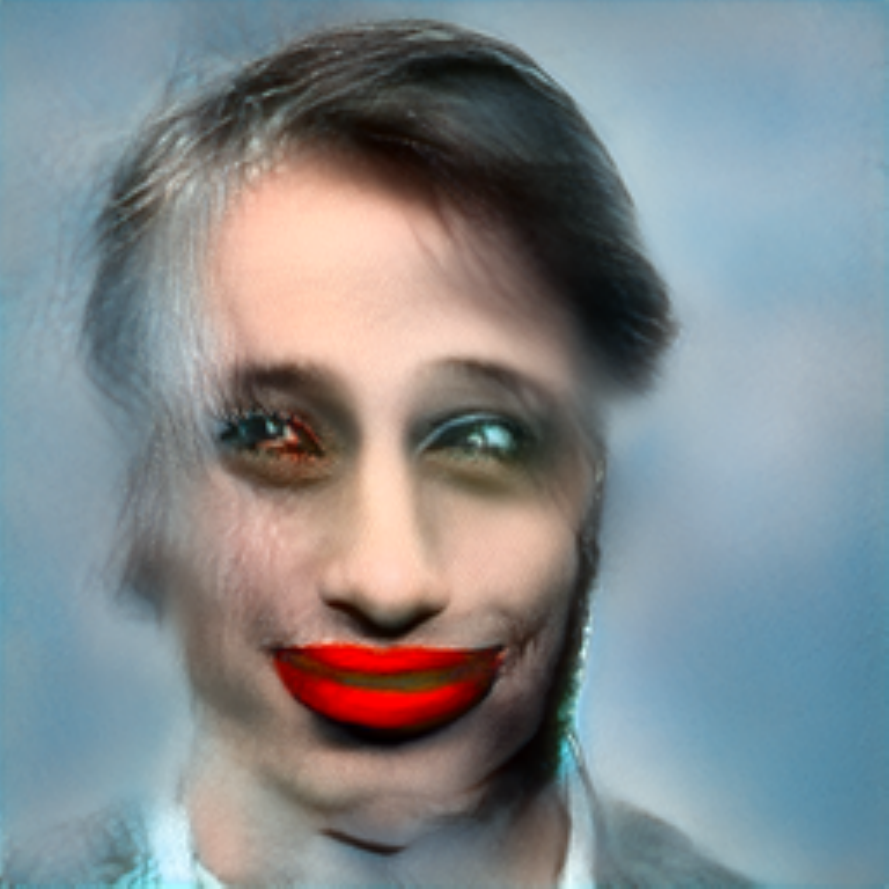} \\
\includegraphics[width=\linewidth]{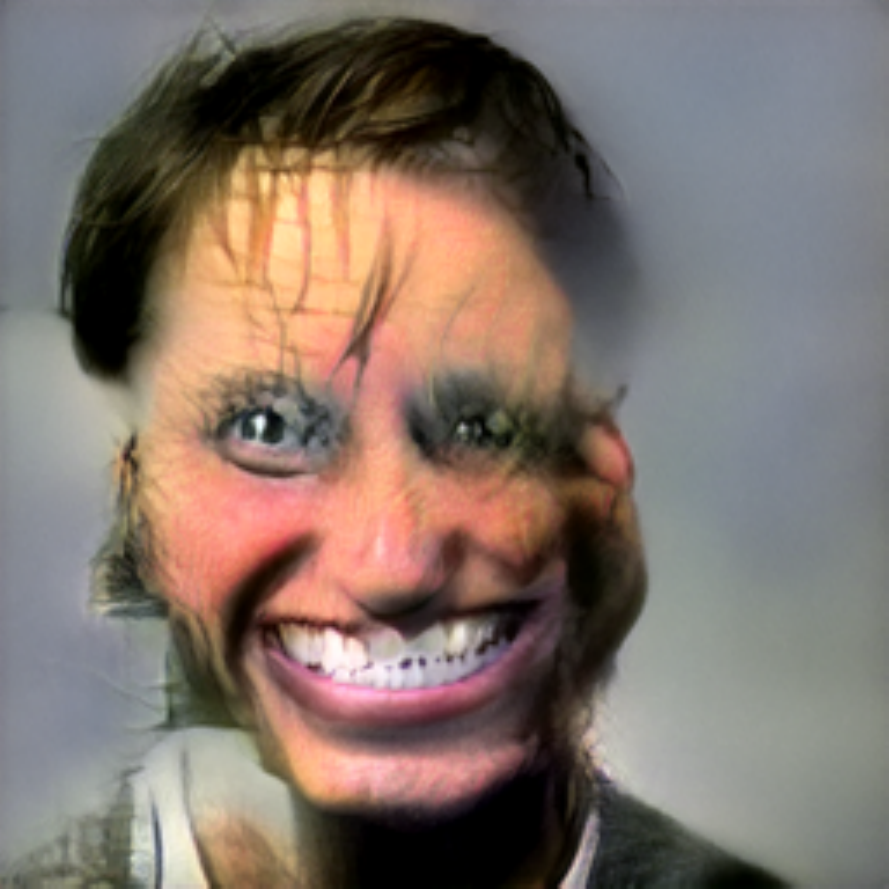} \\
\end{tabular}
\end{minipage}%
\begin{minipage}[b]{0.02\linewidth}
\end{minipage}%
\begin{minipage}[b]{0.16\linewidth}
\begin{tabular}{c}
\footnotesize $\FZS$ \\
\includegraphics[width=\linewidth]{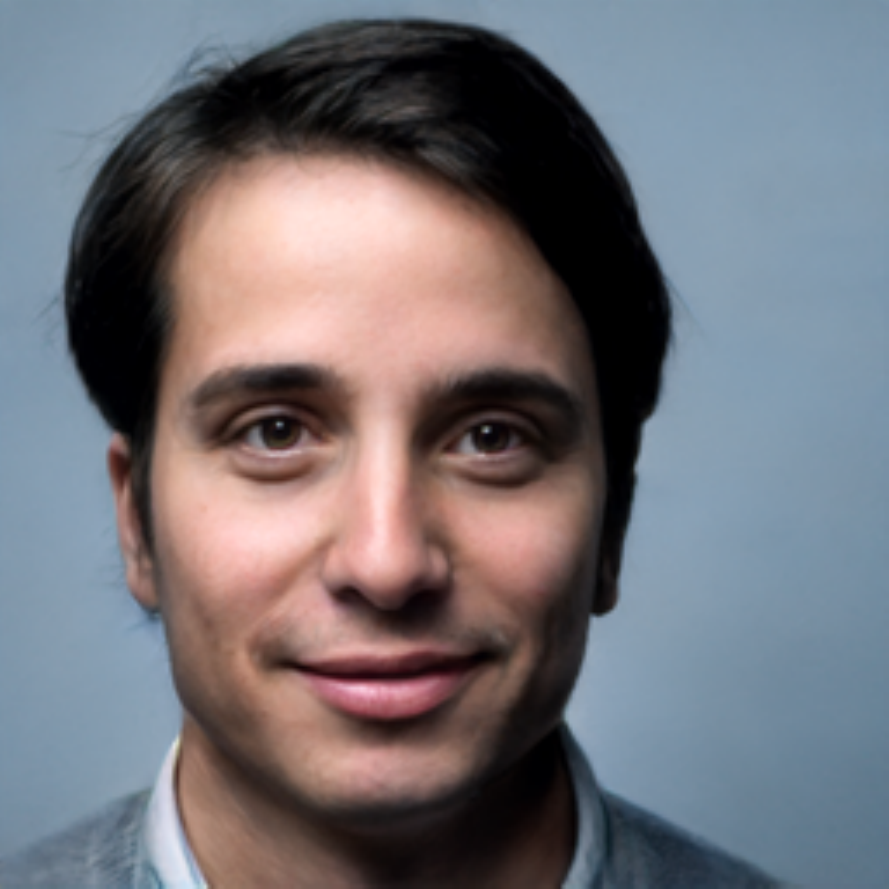} \\
\includegraphics[width=\linewidth]{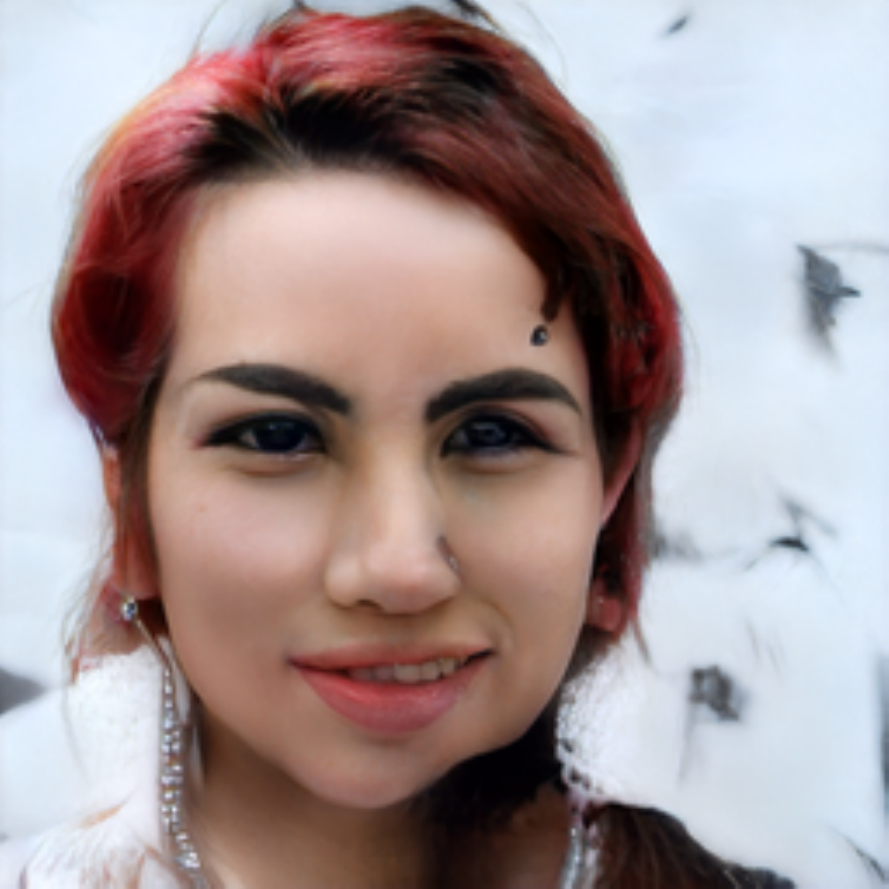} \\
\includegraphics[width=\linewidth]{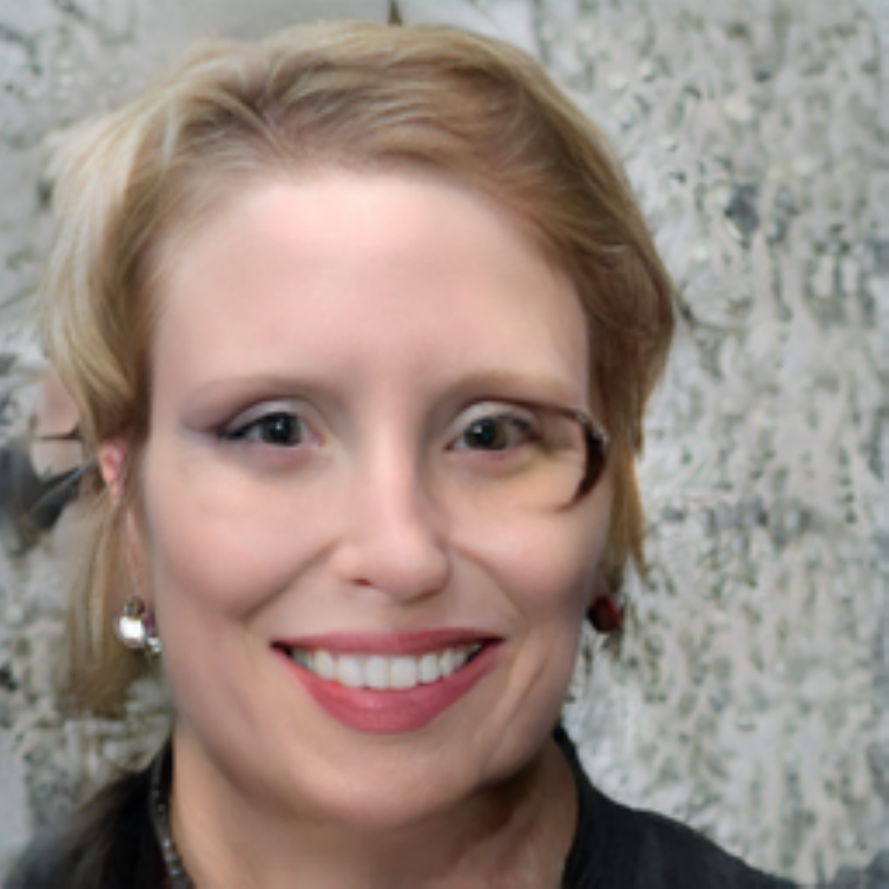} \\
\end{tabular}
\end{minipage}%
\begin{minipage}[b]{0.02\linewidth}
\end{minipage}%
\begin{minipage}[b]{0.16\linewidth}
\begin{tabular}{c}
\footnotesize $\FWS$\!~\!\cite{Kang_2021_ICCV} \\
\includegraphics[width=\linewidth]{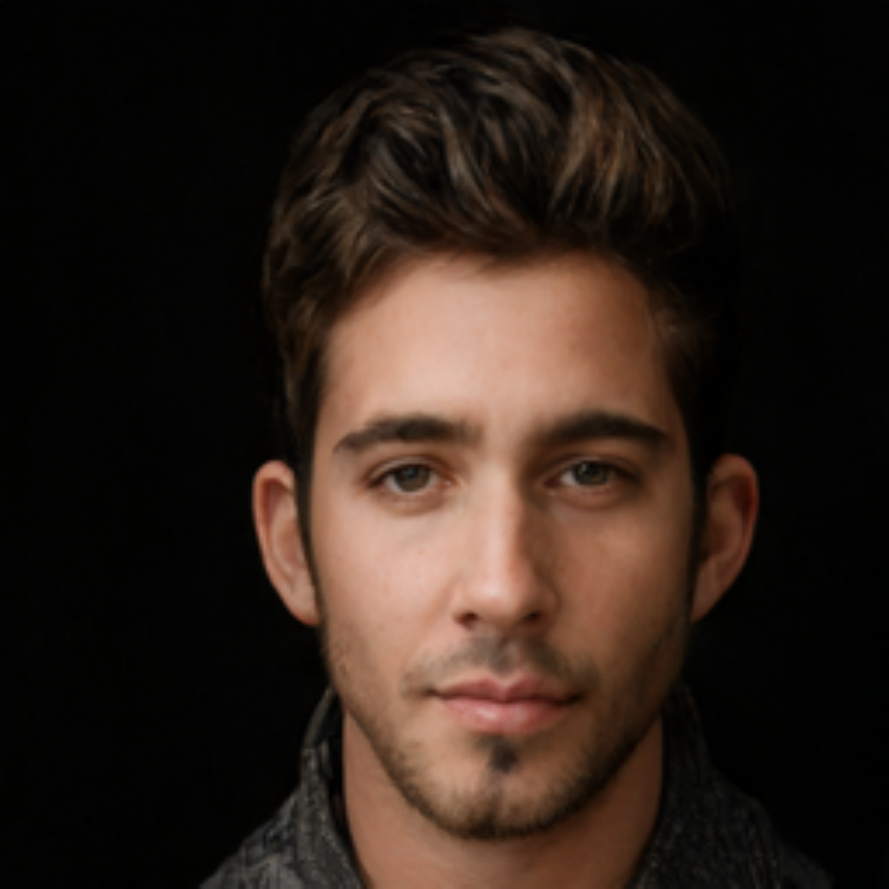} \\
\includegraphics[width=\linewidth]{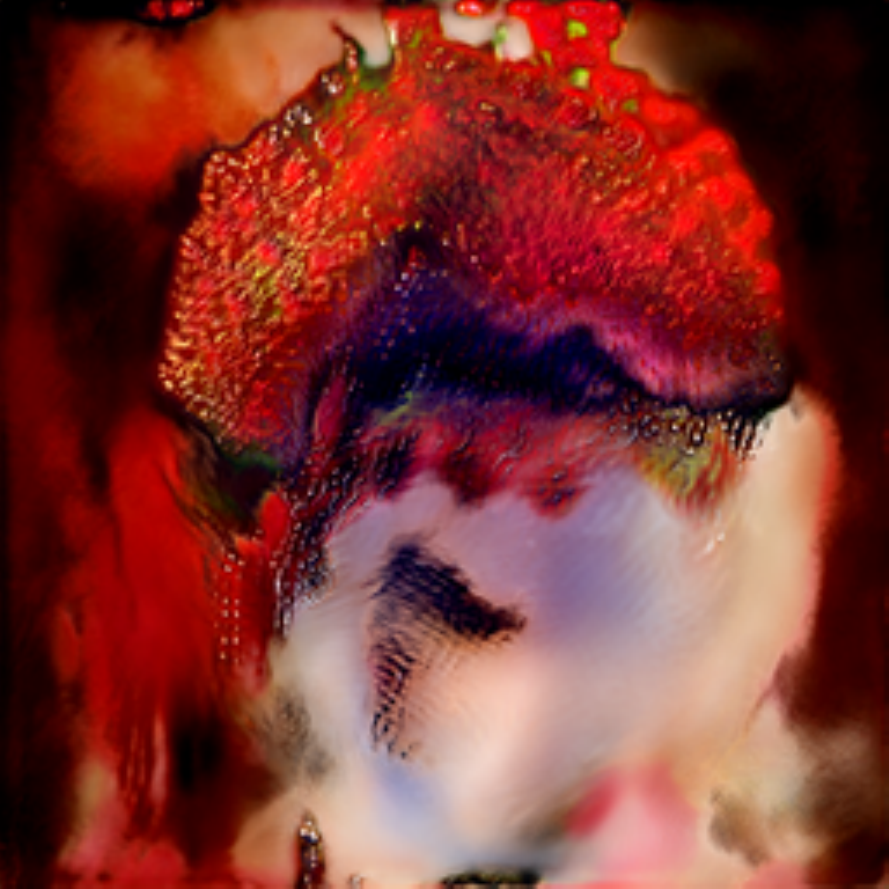} \\
\includegraphics[width=\linewidth]{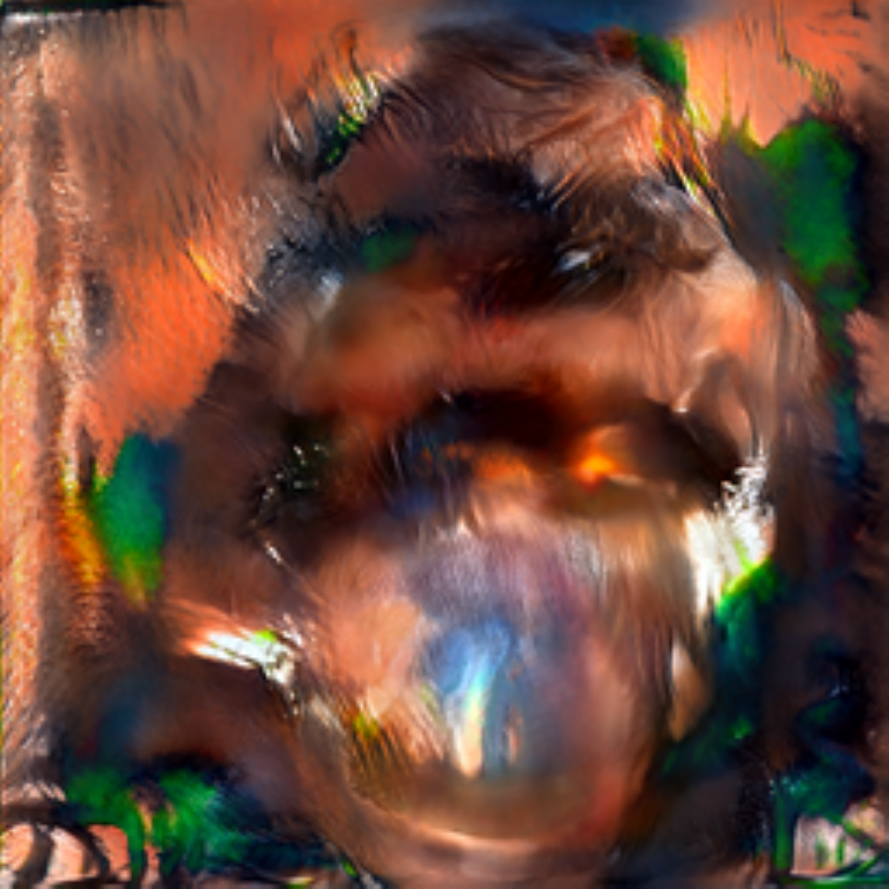} \\
\end{tabular}
\end{minipage}%
\begin{minipage}[b]{0.02\linewidth}
\end{minipage}%
\begin{minipage}[b]{0.16\linewidth}
\begin{tabular}{c}
\footnotesize $\mathcal{F}\!/\!\mathcal{S}$~\cite{yao2022style} \\
\includegraphics[width=\linewidth]{interfacegan/sp_sample3_1.pdf} \\
\includegraphics[width=\linewidth]{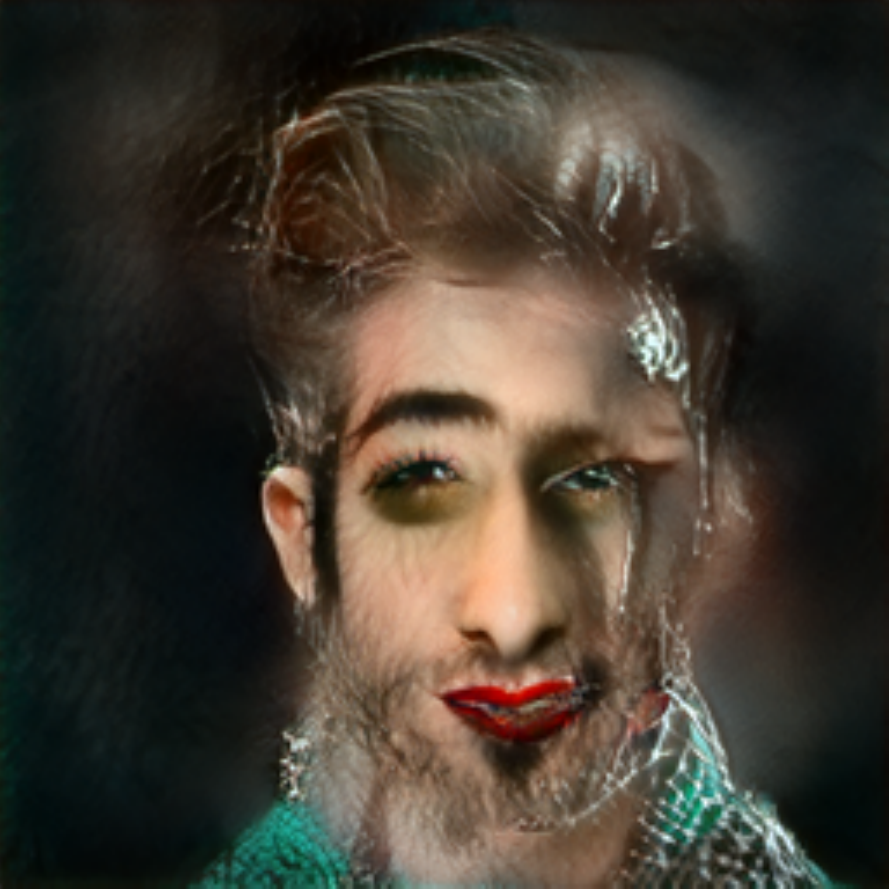} \\
\includegraphics[width=\linewidth]{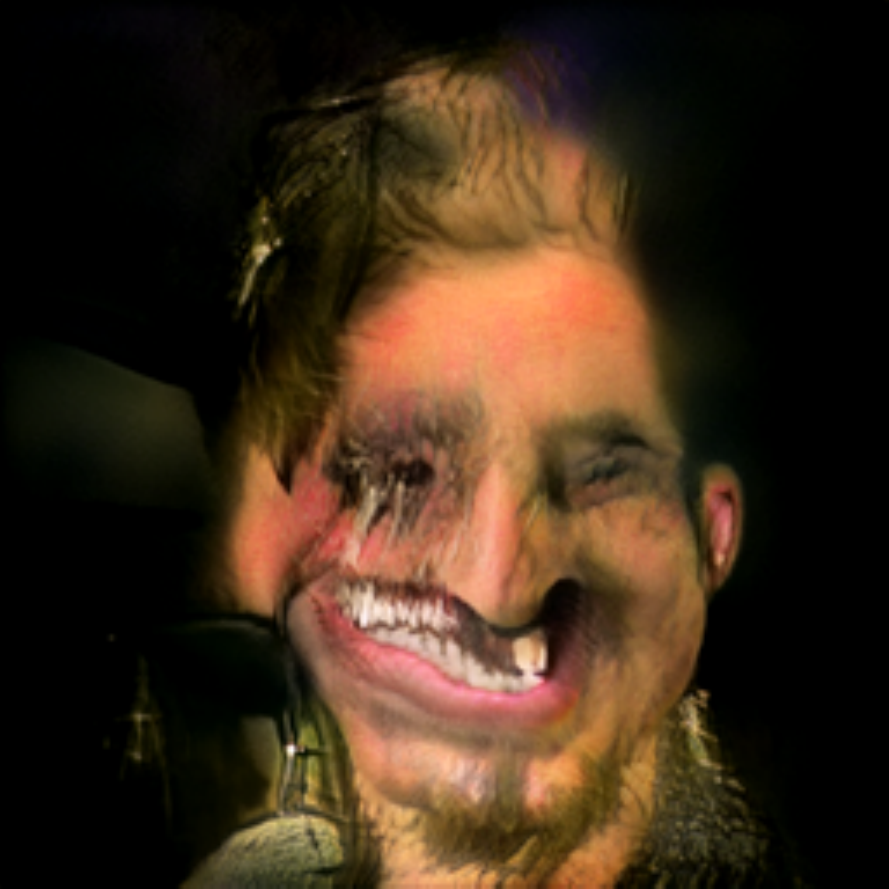} \\
\end{tabular}
\end{minipage}%
\begin{minipage}[b]{0.02\linewidth}
\end{minipage}%
\begin{minipage}[b]{0.16\linewidth}
\begin{tabular}{c}
\footnotesize $\FZS$ \\
\includegraphics[width=\linewidth]{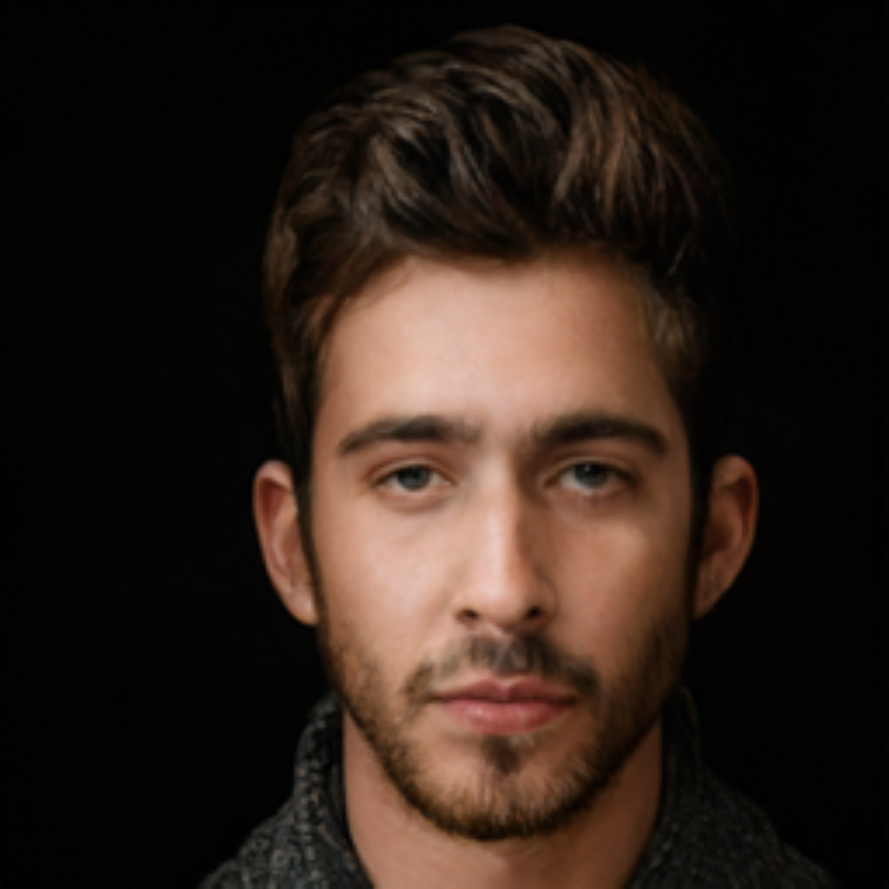} \\
\includegraphics[width=\linewidth]{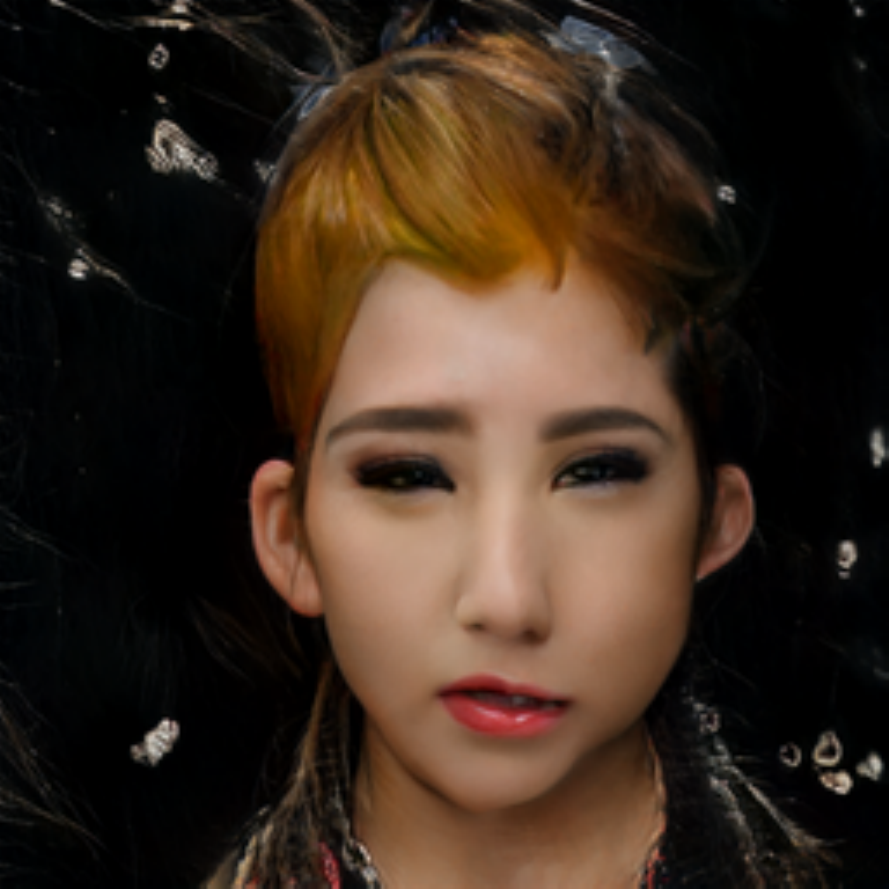} \\
\includegraphics[width=\linewidth]{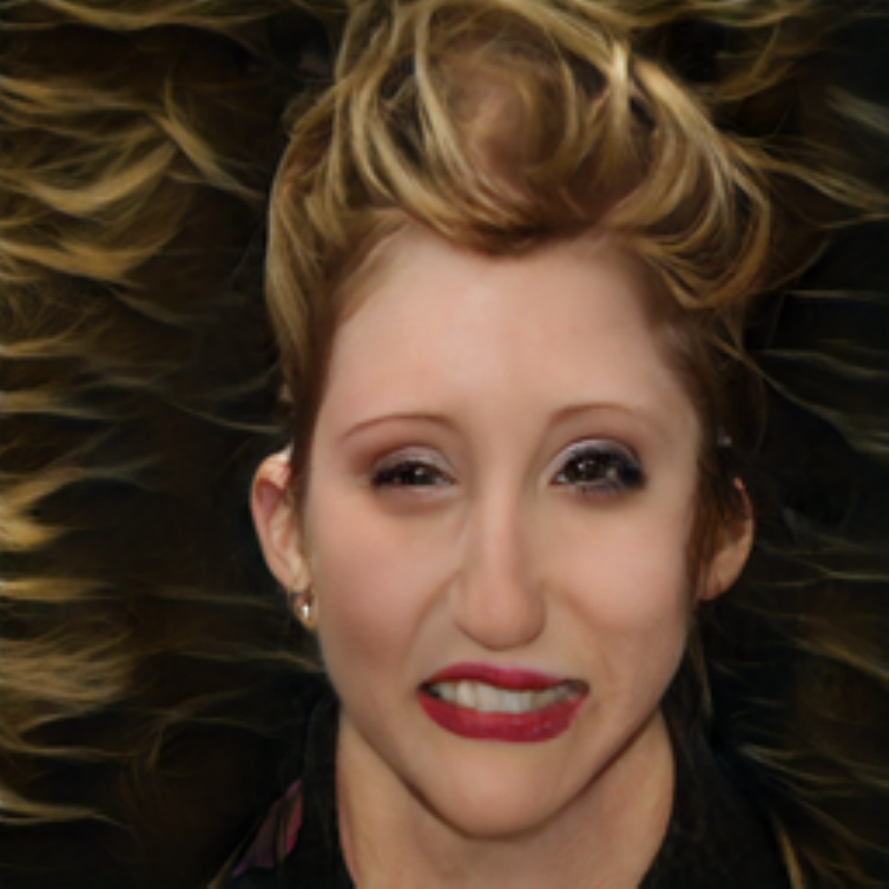} \\
\end{tabular}
\end{minipage}%
\egroup
\vspace{-0.8em}
\caption{Editing comparison with InterfaceGAN~\cite{shen2020interfacegan} directions.\vspace{-0.8em}}\label{fig:interfacegan}
\end{figure}

\begin{figure}[tb]
\centering
\includegraphics[width=0.95\columnwidth]{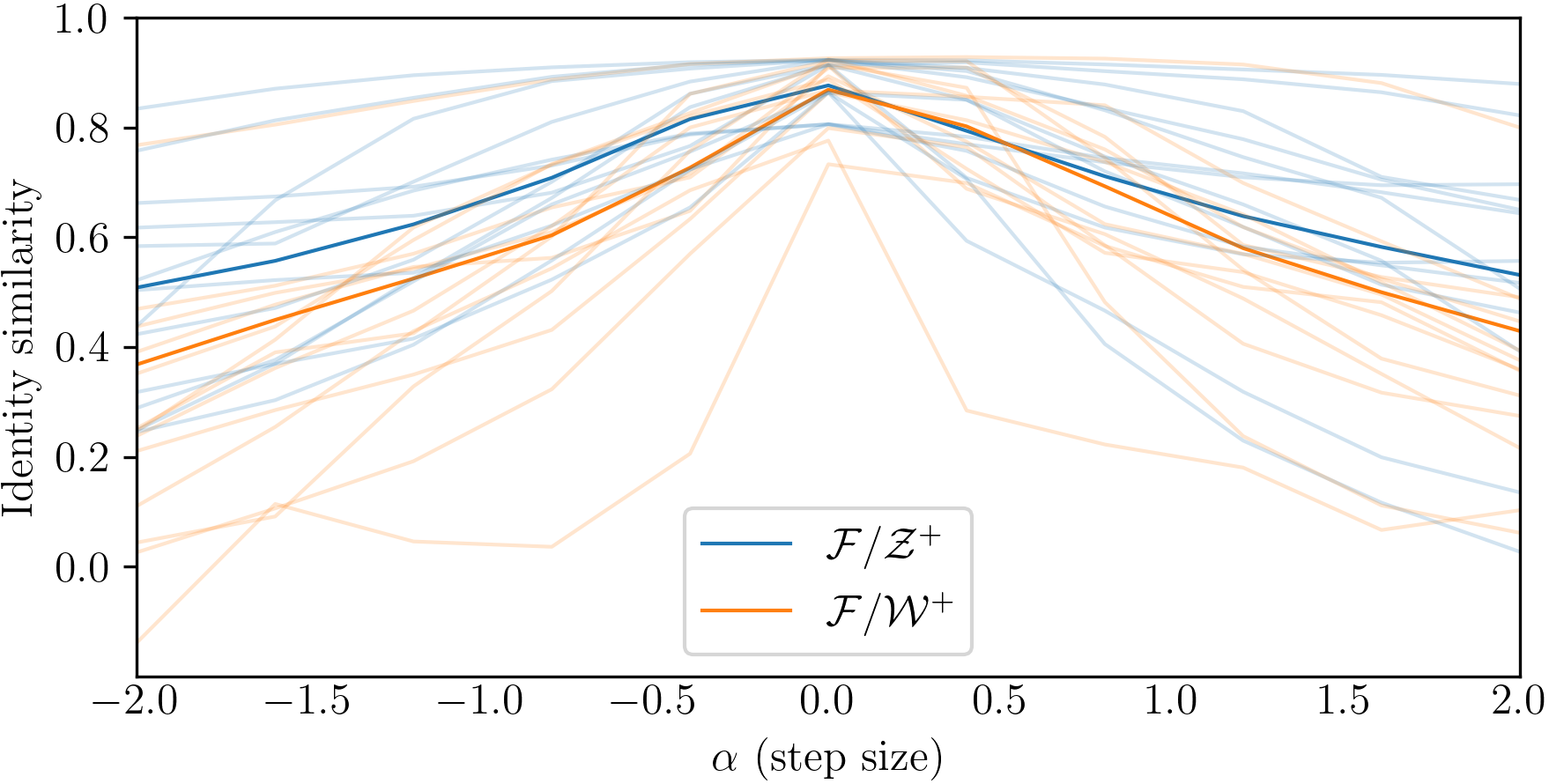}
\vspace{-0.8em}
\caption{Quantitative comparison of the identity similarity between target and edited images. Light-colored lines indicate the individual results. Deep-colored lines indicate the average similarity.
$\FZS$ shows high editing quality.\vspace{-0.8em}}\label{fig:identity_similarity}
\end{figure}

\subsection{Overall Concept}
Overall, there is still no existing latent space that can guarantee both reconstruction quality and editing quality. As discussed in \cite{menon2020pulse,zhu2020improved,song2021agilegan,wulff2020improving}, leveraging $\ZS$ or $\ZPS$ spaces lead to high editing quality in exchange for reconstruction quality.
PULSE~\cite{menon2020pulse} discussed the importance of
considering a manifold of a latent space, which controls content quality. Following this discussion, Zhu \etal\cite{zhu2020improved} uses regularization on the deactivated $\WS$ becase they cannot use $\ZS$ space with sufficient reconstruction quality.
To greatly benefit from considering the latent manifold, we employ bounded latent codes. Since we know the shape of $\ZS$,
we completely utilize the information of the distribution of $\ZS$.
To overcome their limitations, we employ an auxiliary space to improve reconstruction quality while using $\ZPS$.
Since current cutting-edge GAN inversion methods still have over-capacity, we can replace their latent spaces
to $\ZPS$ without sacrifing reconstruction quality.

The space with $\ZPS$ has the desirable properties required for GAN inversion: high reconstruction quality and high editing quality.
The high reconstruction capacity of the space is attributed to the auxiliary space, and the high editing quality is attributed to the primitive space $\ZS$.

\subsection{GAN inversion with \texorpdfstring{$\ZPS$}{Z+} space}

To demonstrate the effectiveness of the use of $\ZPS$,
we replace $\WS$ or $\WPS$ in BDInvert~\cite{Kang_2021_ICCV}, SAM~\cite{parmar2022spatially}, and PTI~\cite{roich2021pivotal}.
BDInvert~\cite{Kang_2021_ICCV} and SAM~\cite{parmar2022spatially} use $\FWS$ space.
PTI~\cite{roich2021pivotal} uses $\WS$ and fine-tunes the generator to enhance reconstruction quality and increase the density of the inverted neighborhood.
For example, for BDInvert~\cite{Kang_2021_ICCV} we replace $\FWS$ with $\ZPS$ and present the $\FZS$ space shown in \cref{fig:fzpspace}.
The space $\FZS$ cosists of $\FS$ and $\ZPS$, and each elements $\fz \in \FZS$ is defined as $\fz = (\f, \zmpls)$, where $\zmpls = \{\z_i\}_{i=M}^{N}$.
To optimize a latent code along $\ZPS$ or $\ZS$,
we retract the latent codes to the surface of the hypersphere of radius $\sqrt{512}$ each iteration by using:
\begin{align}
  \z_i = \sqrt{512}\z_i/|\z_i|,
\end{align}
where $\z_i \in \zmpls$. We follow the optimization algorithms of the base methods (\ie, BDInvert, SAM, and PTI).

\section{Experiments}
We evaluate the latent spaces from two aspects: reconstruction quality and editing quality.
For the reconstruction quality comparison, we verify that our space is not inferior to the compared spaces.
For the comparison of editing quality, we show that our space preserves the perceptual quality of edited images better than the other spaces.

\noindent \textbf{Reconstruction quality comparison.}
We first compare the reconstruction performance 
using a StyleGAN2 model pre-trained on FFHQ~\cite{Karras2019style}. 
\Cref{fig:recon} shows the reconstructed results and LPIPS loss for five benchmark images on the four compared latent spaces. 
All methods reconstruct them well because $\FS$ magnifies the capacity of the latent space.
The LPIPS loss of our $\FZS$ space is comparable to that of $\FWS$.
\Cref{fig:recon} shows that the inverted images of $\FZS$ and $\FWS$ are almost the same as the target images.

For quantitative comparison,
\cref{tb:recon} reports the average of MSE loss and SSIM over 50 random images from CelebA-HQ~\cite{karras2018progressive}.
The results of $\FZS$ are competitive with those of $\FWS(P_N)$, as shown by a non-inferiority test with a margin of $1\times10^{-8}$, yielding p-values of .002983 for MSE and .001609 for SSIM.
The $\FWS(P_N)$ space, however, results in less realistic edited images as seen later.

\begin{figure}[tb]
  \centering
    \bgroup 
    \def\arraystretch{0.2} 
    \setlength\tabcolsep{1pt}
    \begin{tabular}{cccccc}
& inversion & edited & &inversion& edited \\
\raisebox{1.0em}{\shortstack{\small $\FWS$\\ ($\PNS$)\cite{Kang_2021_ICCV}}}
&\includegraphics[width=0.16\columnwidth]{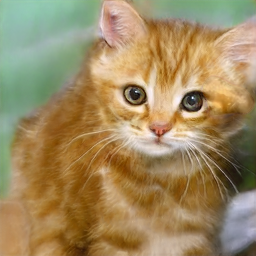} &
\includegraphics[width=0.16\columnwidth]{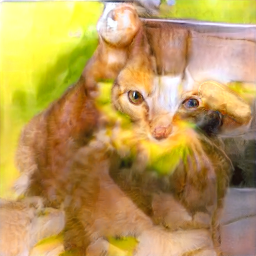} &
\raisebox{1.0em}{\shortstack{\small $\mathcal{F}/\mathcal{Z}^{+}$ \\ (Ours)}} & 
\includegraphics[width=0.16\columnwidth]{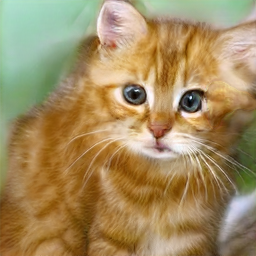} &
\includegraphics[width=0.16\columnwidth]{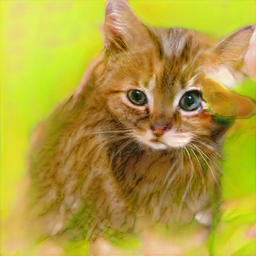} \\
\end{tabular}\egroup \vspace{-0.8em}
    \caption{Edited results on the LSUN Cat dataset with StyleGAN1. Editing on our $\FZS$ space
      preserves the image content (cat).\vspace{-2em}
    }\label{fig:cat_stylegan1}
\end{figure}

\noindent \textbf{Editing quality comparison.}
\Cref{fig:editing} shows editing results with 
GANSpace~\cite{NEURIPS2020_ganspace}. 
For each direction, two images with intensities of -2 and 2 are plotted.
$\FWS(P_N)$ and $\FWS$ lack the image quality after performing editing operation (\eg, lacking face parts or are adding waterdrops).
Meanwhile, our $\FZS$ and $\FZ$ spaces 
consistently preserve image quality after semantic editing.
We also compare editing results with interfaceGAN. As shown in \cref{fig:interfacegan}, we can see that our method relaxes
the distortions of edited images more than the competing methods.

Finally, we evaluate the editing quality quantitatively.
We use MTCNN~\cite{zhang2016joint} as face detector and InceptionResNet V1~\cite{szegedy2017inception} trained on VGGFace2~\cite{cao2018vggface2} as feature extractor.
To compute identity similarity, we use cosine similarity. For each method, we plot the identity similarities between the original inputs and the edited images with each editing step size in \cref{fig:identity_similarity}. 
We plot 12 lines for each method (four targets $\times$ three directions).
The figure shows that our space preserves the identity of target images after editing with even a strong intensity unlike $\FWS(\PNS)$.
We also conduct the editing quality evaluation on 50 CelebA-HQ samples with five directions and eleven step sizes. The average identity similarity of $\FZS$ is 0.373, whereas that of $\FWS$ is 0.327.
It demonstrates that $\FZS$ maintains the perceptual quality of edited image well.

\noindent \textbf{Editing comparison on another dataset.}
We evaluate the effectiveness of $\FZS$ on another GAN model.
\Cref{fig:cat_stylegan1} shows the edited results with StyleGAN 1 pretrained on LSUN Cat.
Although $\FWS(\PNS)$ completely corrupts the cat's face in the edited image, our method maintains it.

\begin{figure}[t]
  \centering
    \bgroup 
    \def\arraystretch{0.2} 
    \setlength\tabcolsep{0.8pt}
    \begin{tabular}{cccccc}
\raisebox{1.5em}{\footnotesize PTI\cite{roich2021pivotal}} &
\includegraphics[width=0.154\columnwidth]{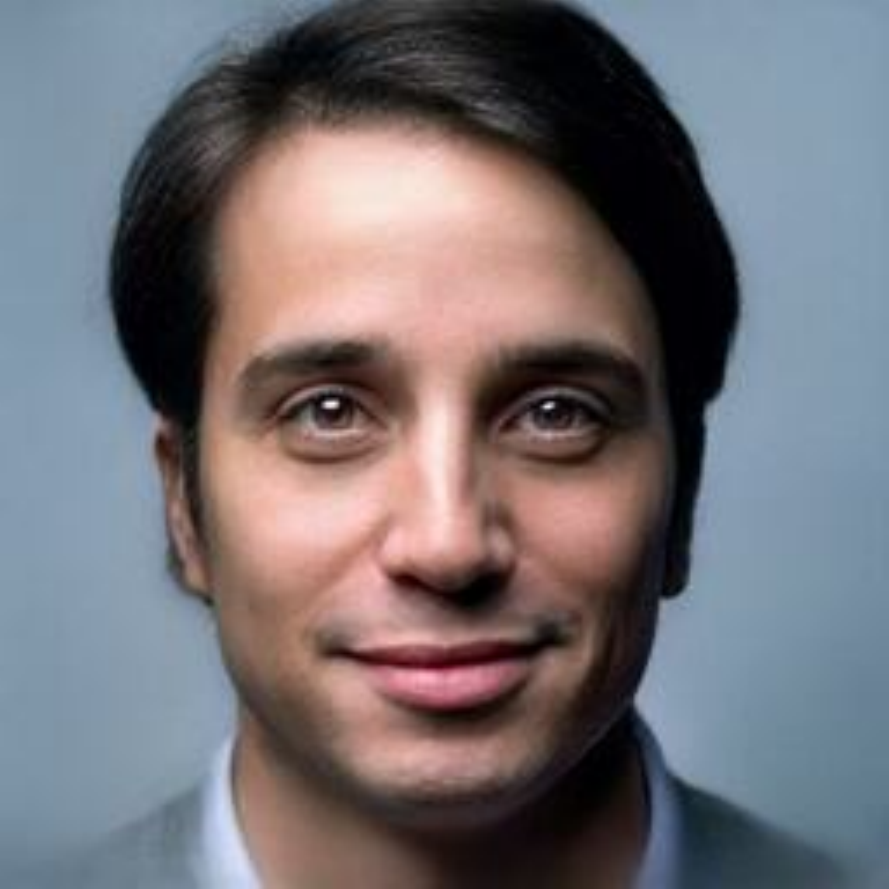} &
\includegraphics[width=0.154\columnwidth]{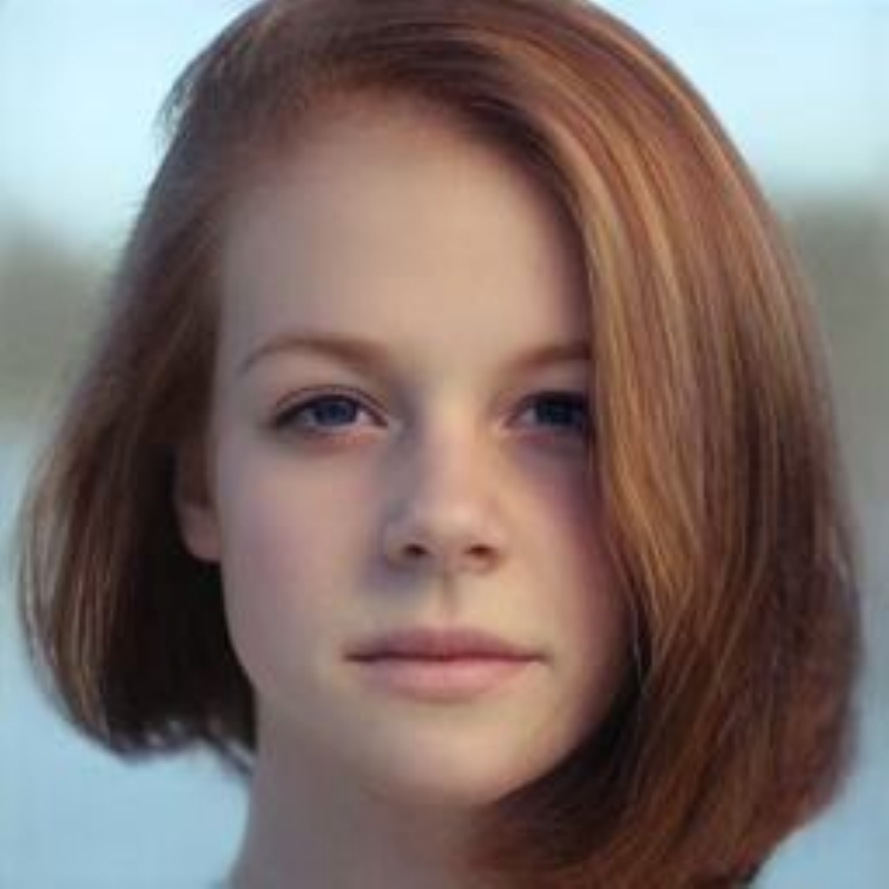} &
\includegraphics[width=0.154\columnwidth]{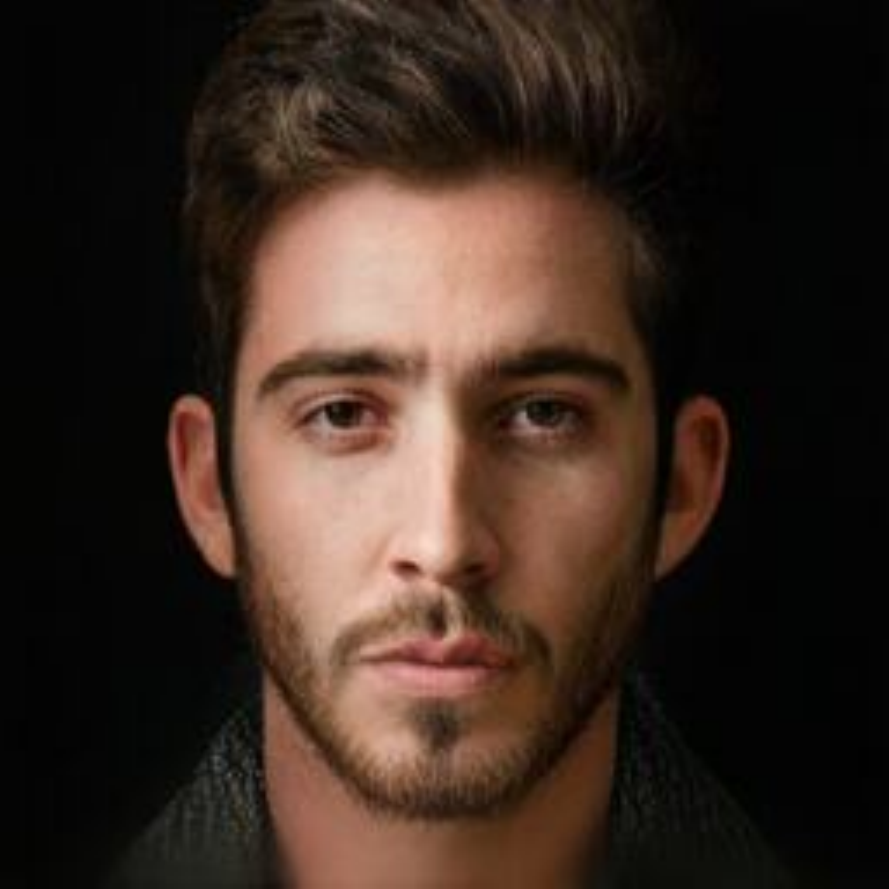} &
\includegraphics[width=0.154\columnwidth]{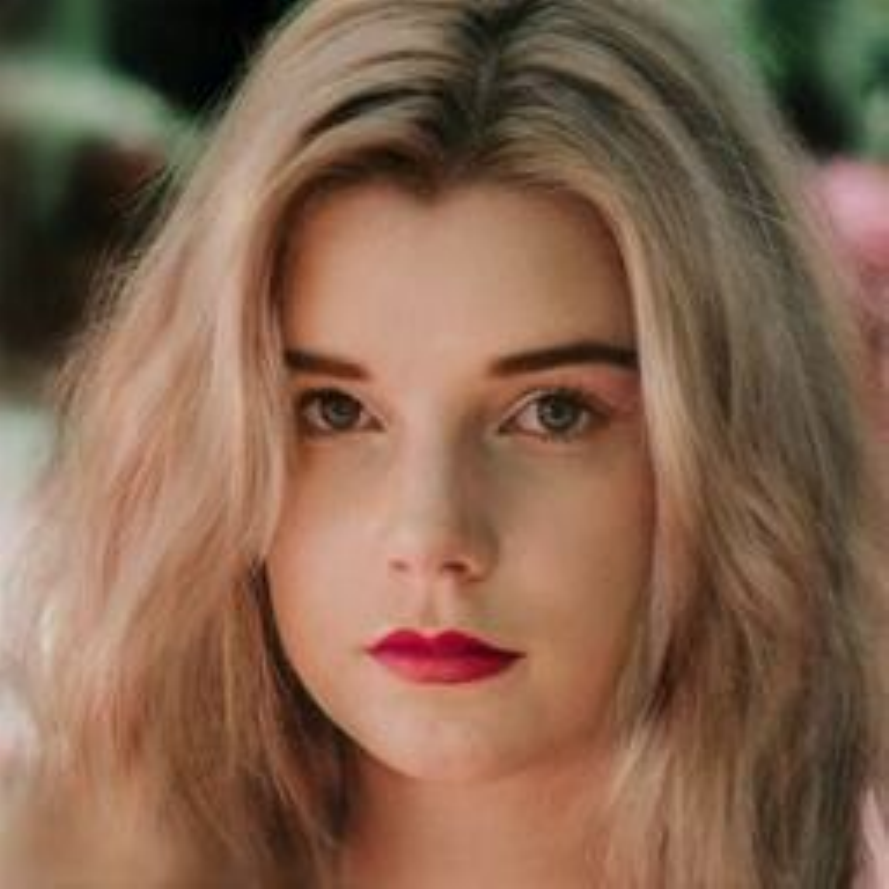} &
\includegraphics[width=0.154\columnwidth]{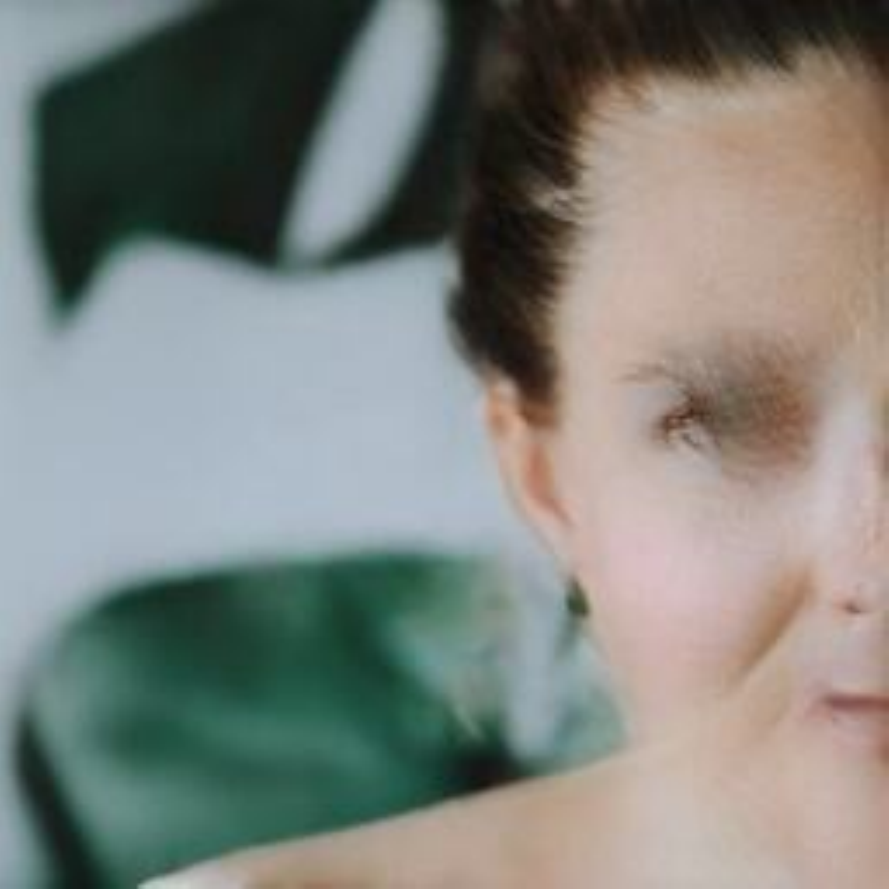} \\
\footnotesize LPIPS & \footnotesize $0.1249$ & \footnotesize $0.1083$ & \footnotesize $0.3031$  & \scriptsize $0.1249$ &  \footnotesize $0.0638$  \\
\raisebox{1.2em}{\footnotesize \shortstack{PTI\cite{roich2021pivotal} \\ \!w/\! $\ZPS$}} &
\includegraphics[width=0.154\columnwidth]{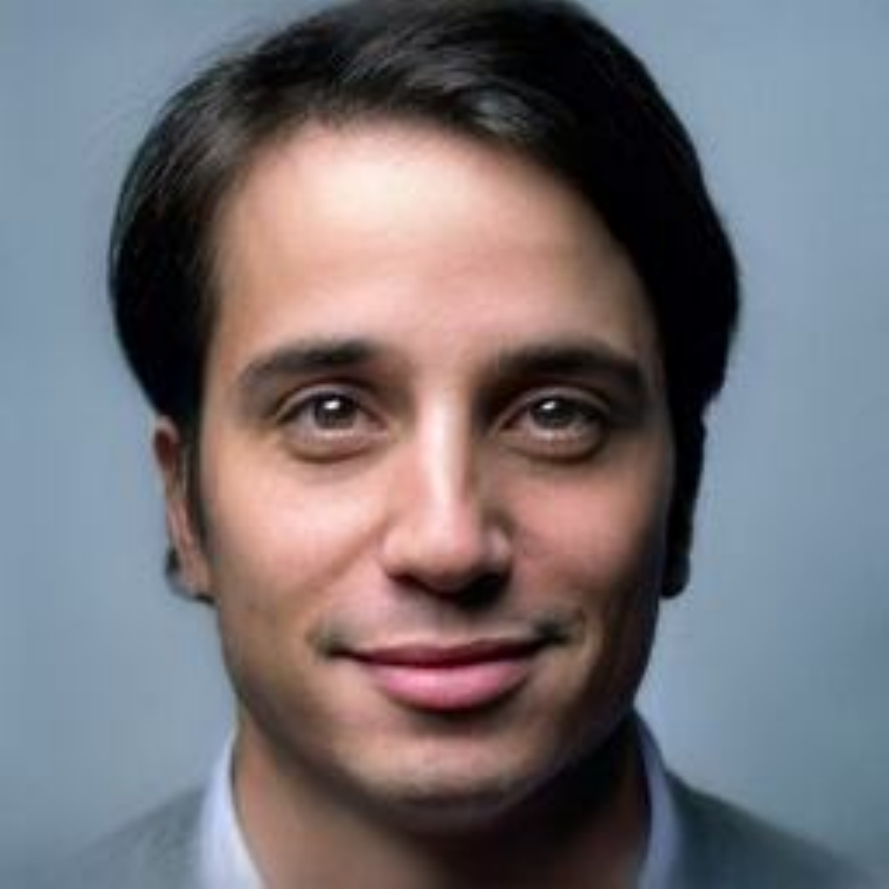} &
\includegraphics[width=0.154\columnwidth]{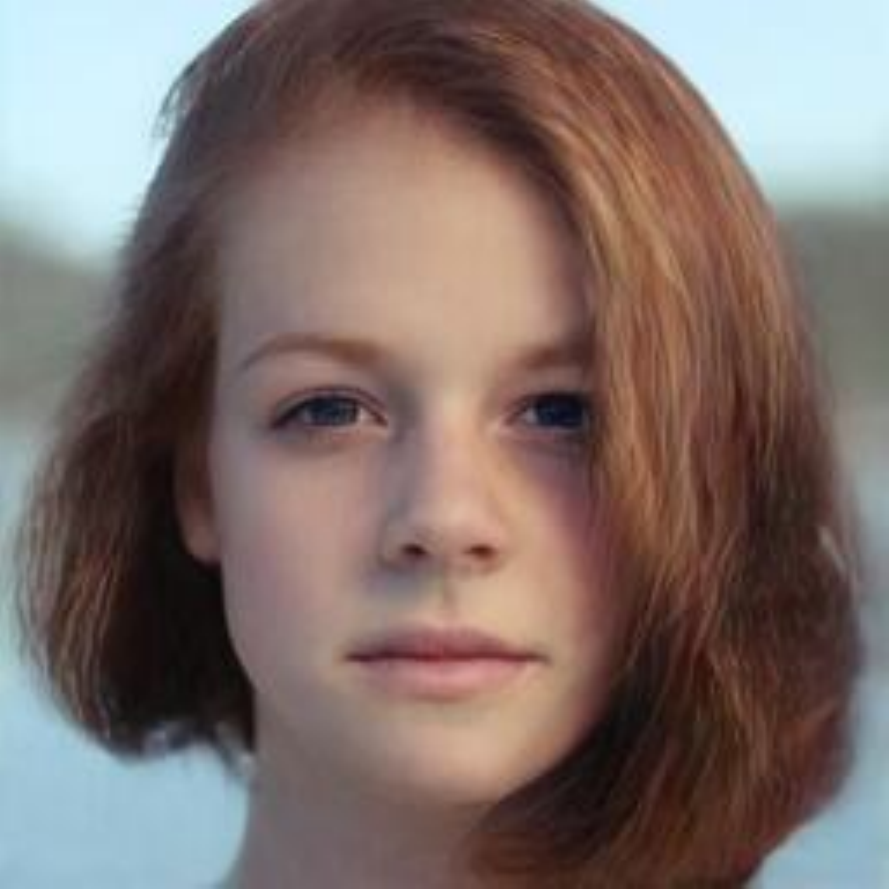} &
\includegraphics[width=0.154\columnwidth]{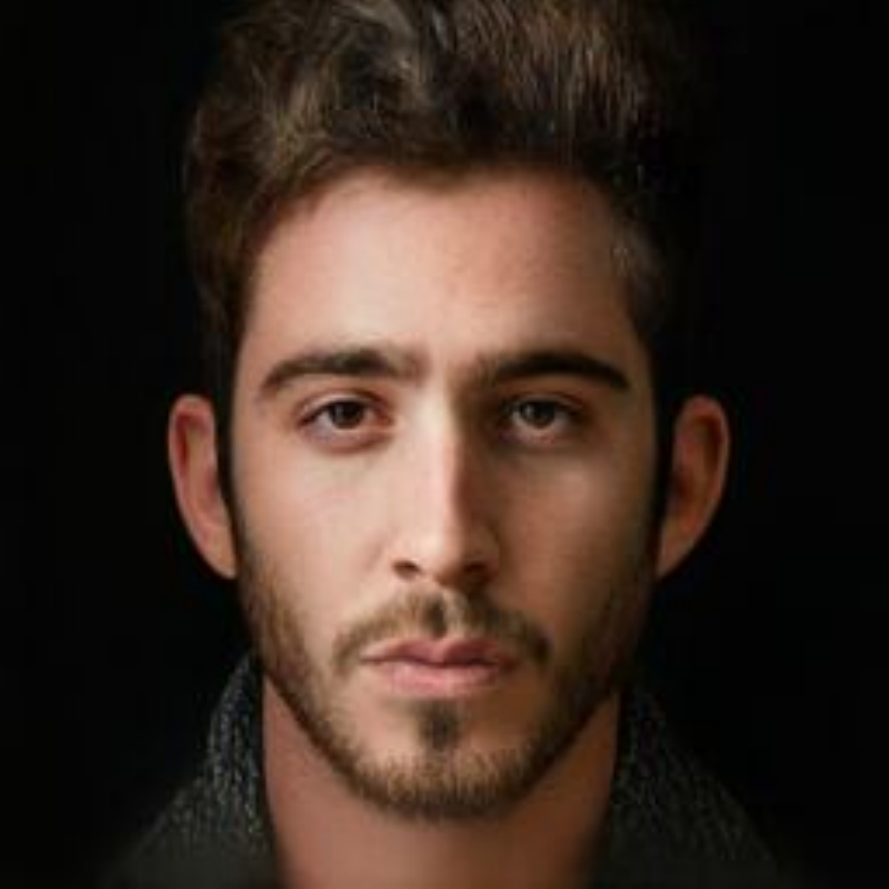} &
\includegraphics[width=0.154\columnwidth]{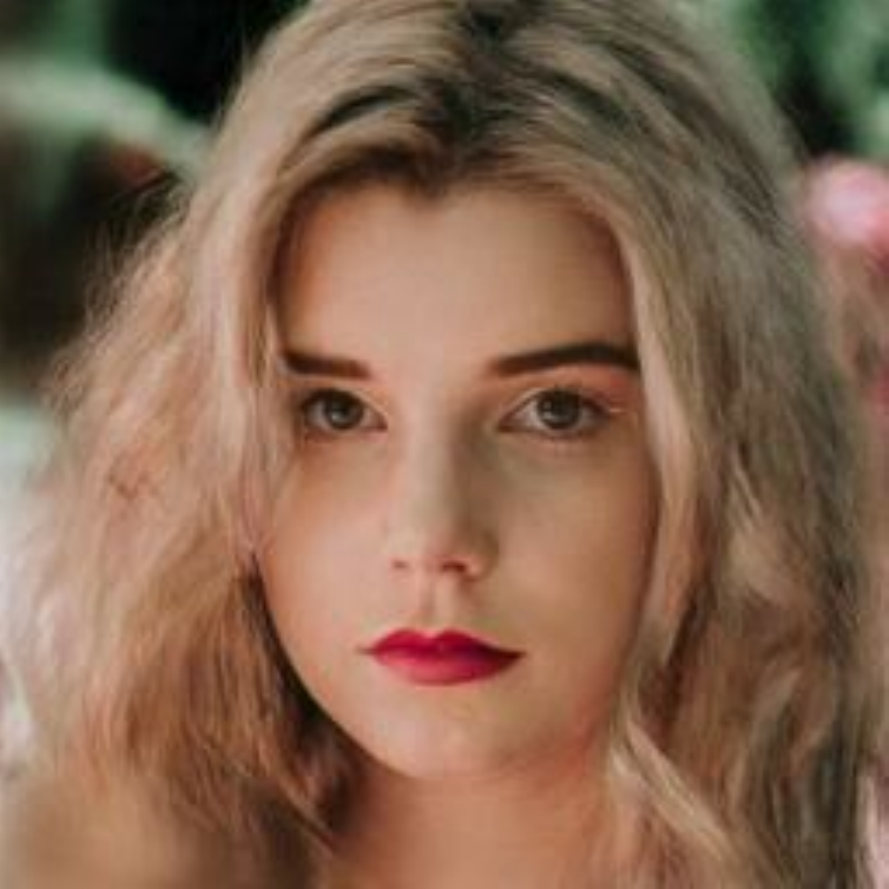} &
\includegraphics[width=0.154\columnwidth]{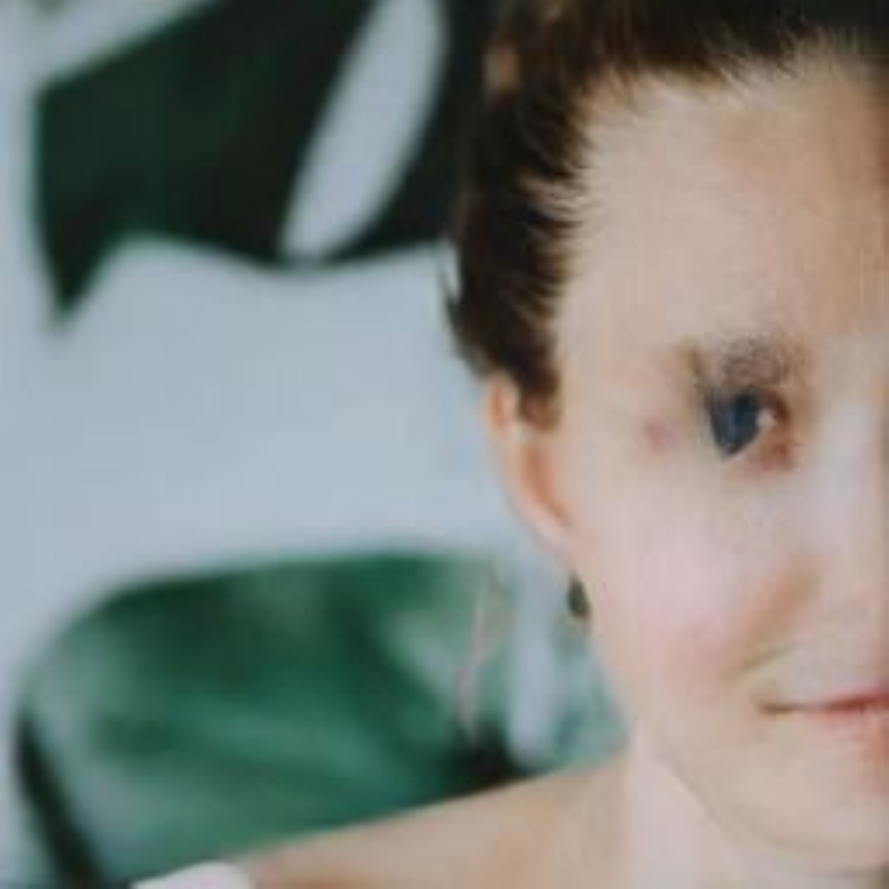} \\
\footnotesize LPIPS & \footnotesize $0.1095$ & \footnotesize $0.0852$ &  \footnotesize $0.3017$ & \footnotesize $0.1230$ & \footnotesize $0.0626$  \\
\raisebox{1.5em}{\footnotesize SAM\cite{parmar2022spatially}} &
\includegraphics[width=0.154\columnwidth]{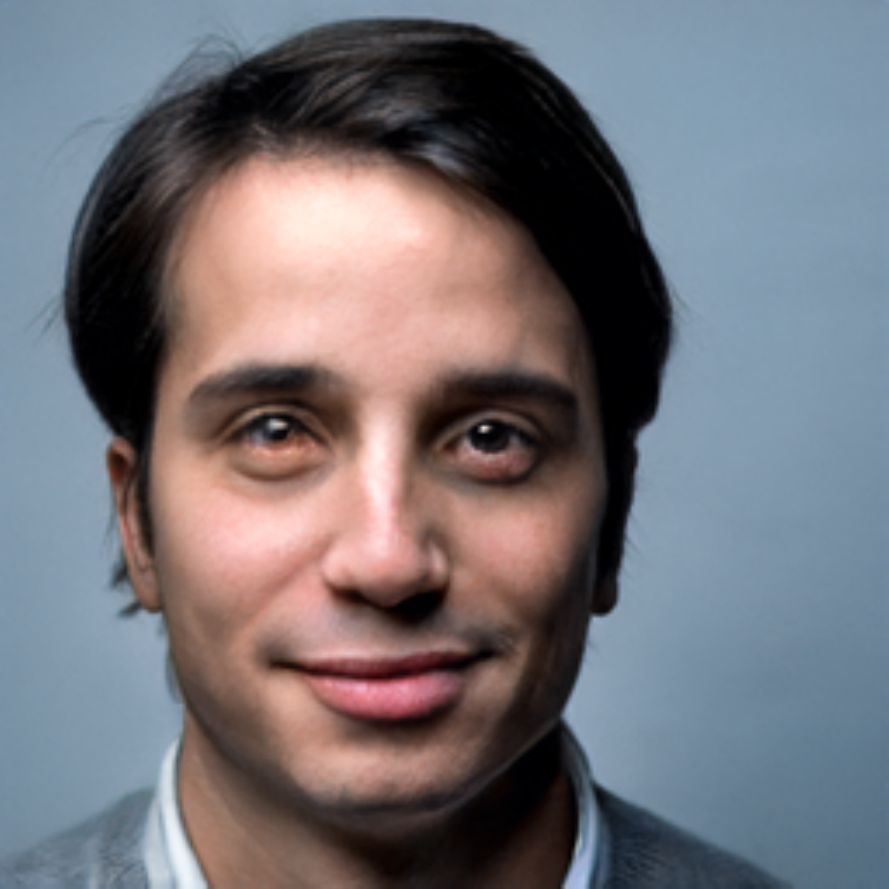} &
\includegraphics[width=0.154\columnwidth]{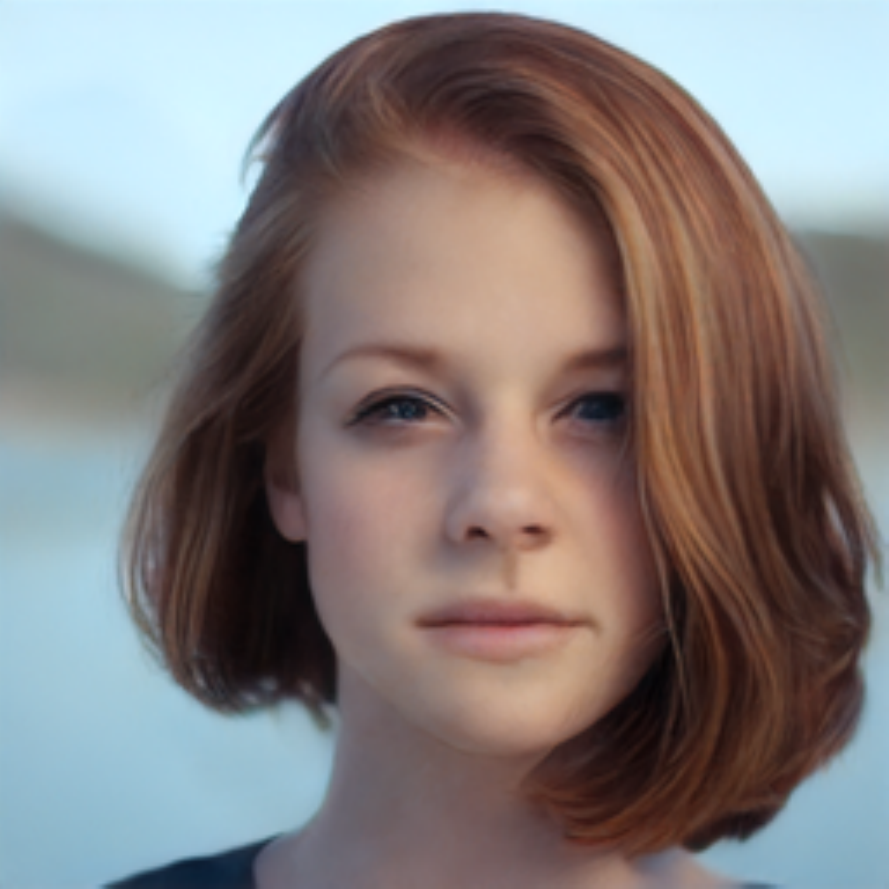} &
\includegraphics[width=0.154\columnwidth]{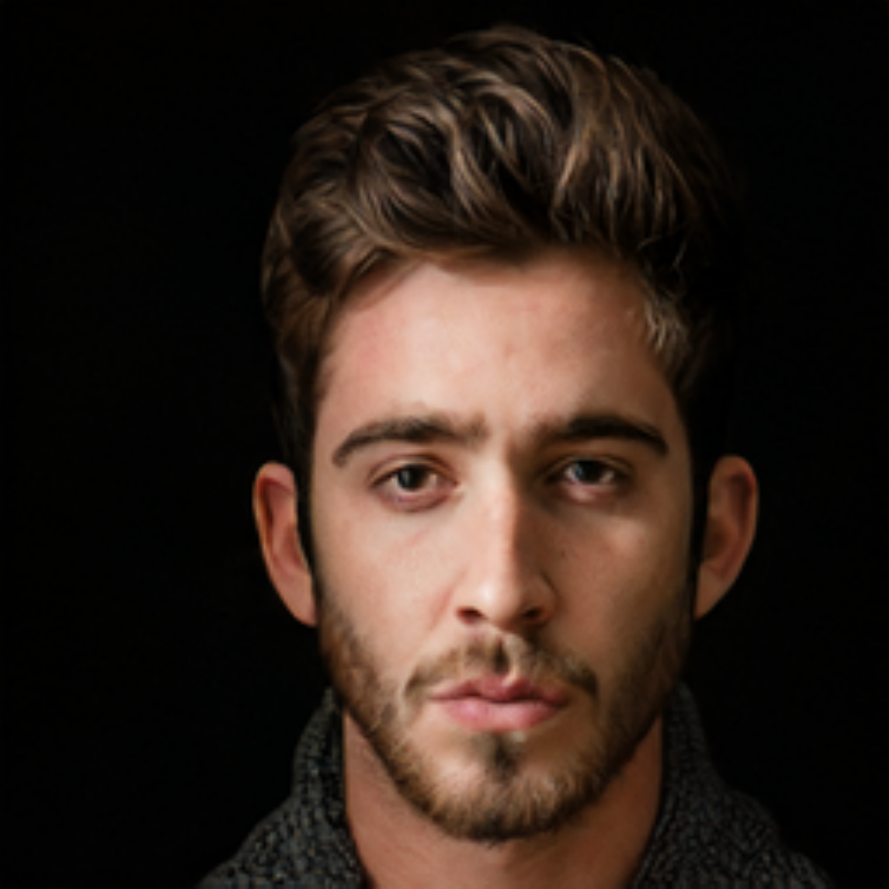} &
\includegraphics[width=0.154\columnwidth]{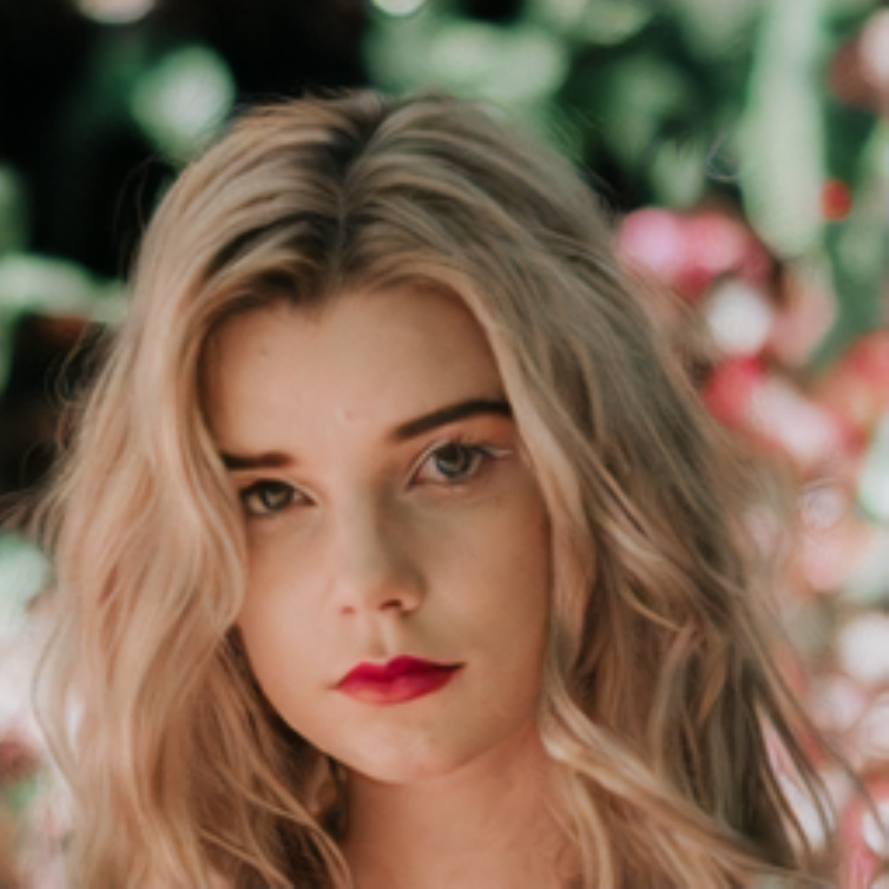} &
\includegraphics[width=0.154\columnwidth]{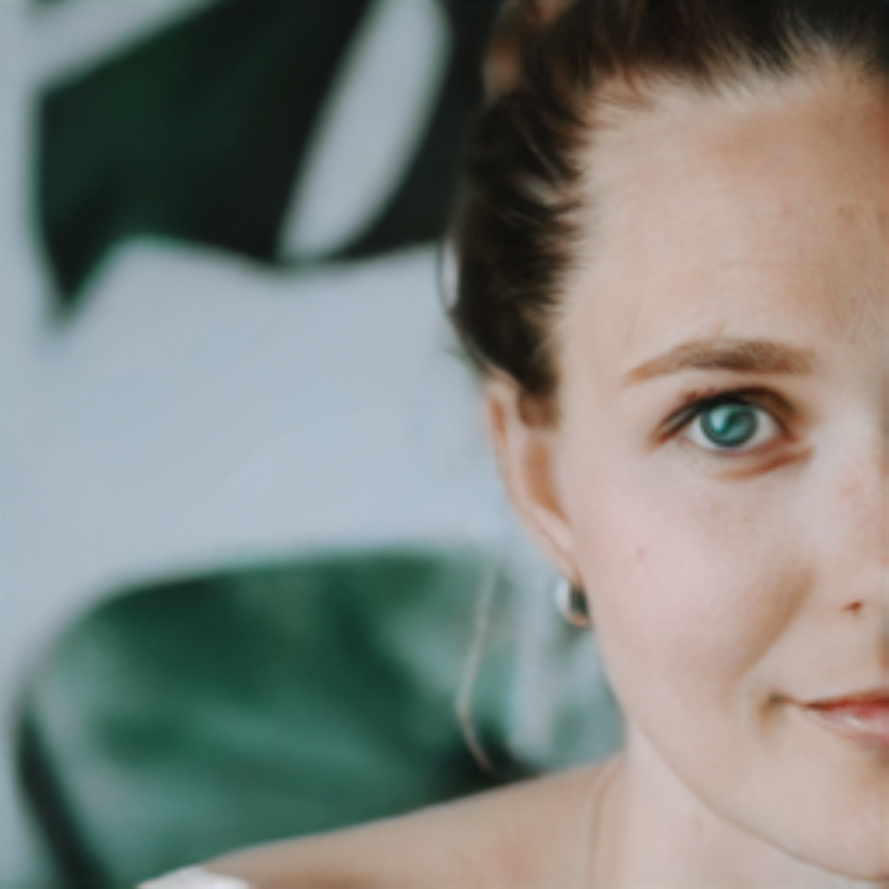} \\
\footnotesize LPIPS & \footnotesize $0.2111$ & \footnotesize $0.1169$ & \footnotesize $0.3802$ & \footnotesize $0.0809$ & \footnotesize $0.0609$ \\
\raisebox{1.2em}{\footnotesize \shortstack{SAM\cite{parmar2022spatially} \\ \!w/\! $\ZPS$}}  &
\includegraphics[width=0.154\columnwidth]{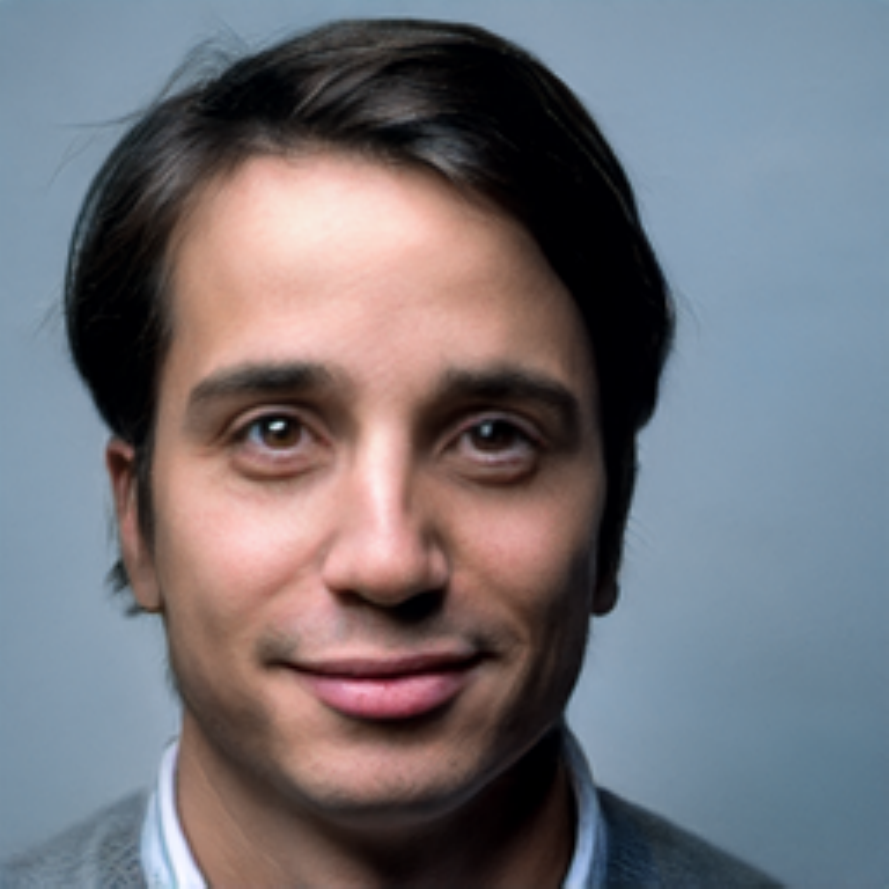} &
\includegraphics[width=0.154\columnwidth]{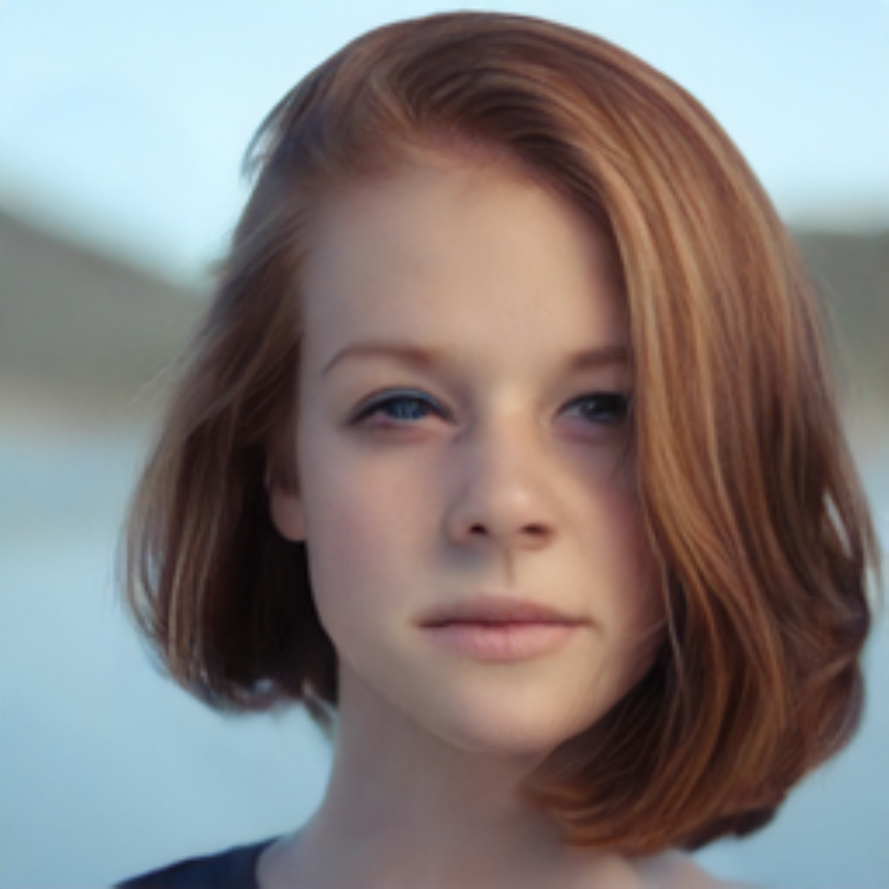} &
\includegraphics[width=0.154\columnwidth]{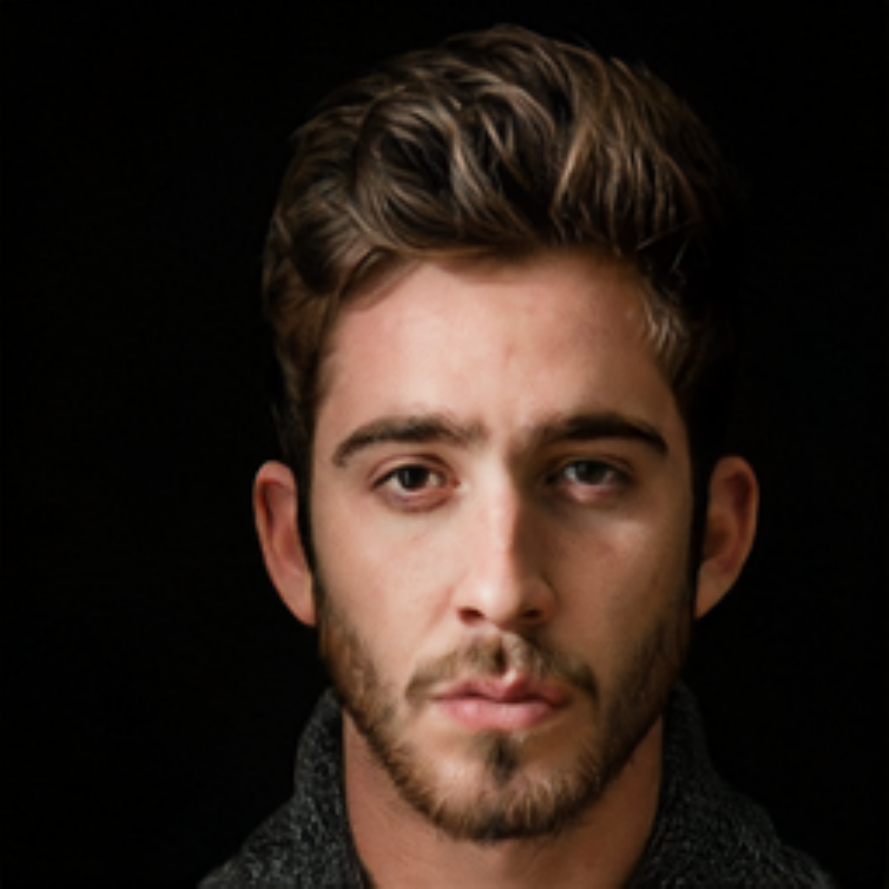} &
\includegraphics[width=0.154\columnwidth]{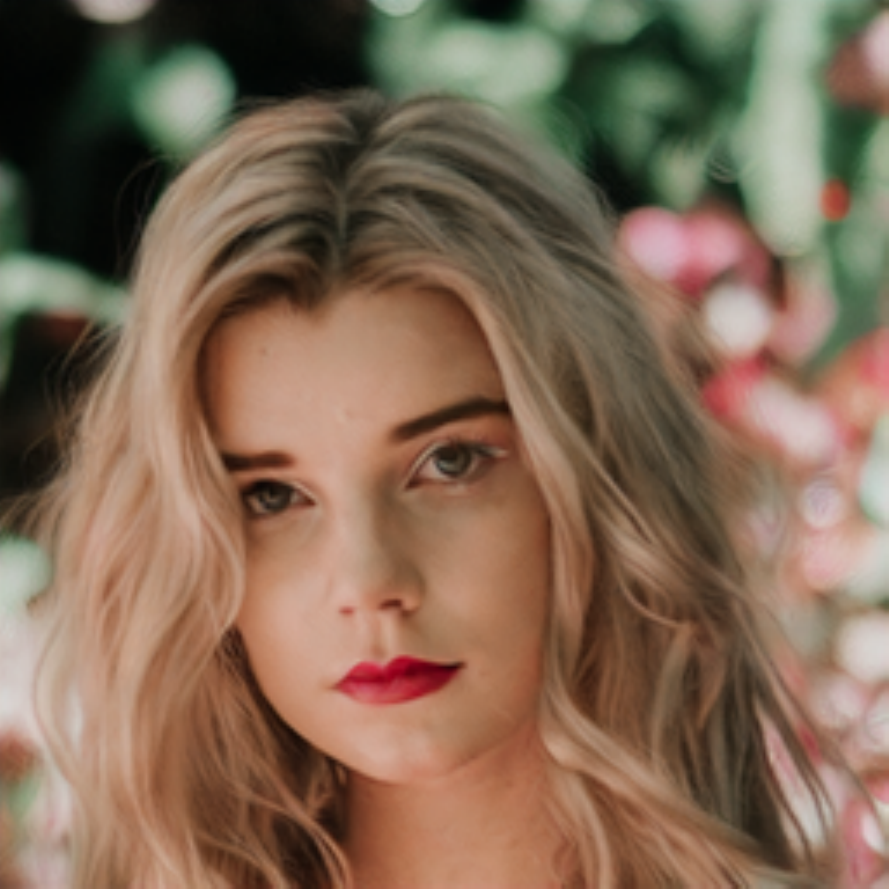} &
\includegraphics[width=0.154\columnwidth]{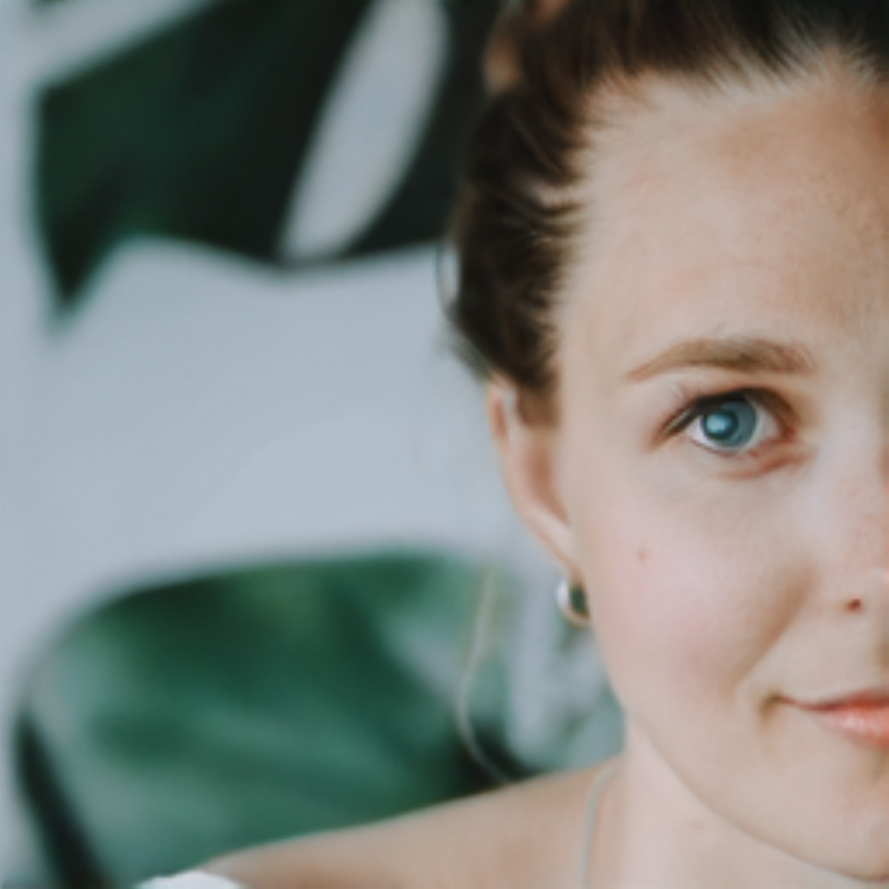} \\
\footnotesize LPIPS &  \footnotesize $0.1987$ & \footnotesize $0.1247$  & \footnotesize $0.3838$ & \footnotesize $0.0793$ & \footnotesize $0.0584$ \\
    \end{tabular}\egroup
\vspace{-0.4em}
\caption{Reconstruction comparisons on SAM and PTI. The 1st and 3rd rows are inverted results of PTI and SAM. The 2nd and 4th rows are results of the methods that use $\ZPS$ instead of $\WS$ or $\WPS$. It indicates we can replace original latent space to $\ZPS$ without loosing reconstruction quality to improve editing quality.\vspace{-0.4em}}\label{fig:sota}
\end{figure}

\begin{figure}[t]
  \centering
    \bgroup 
    \def\arraystretch{0.2} 
    \setlength\tabcolsep{1pt}
    \begin{tabular}{ccccccccc}
\footnotesize inversion & \footnotesize  eyeglass & \footnotesize  smile & \footnotesize \phantom{0} & \footnotesize  inversion & \footnotesize  eyeglass & \footnotesize  smile \\
\includegraphics[width=0.15\columnwidth]{rebuttal/sam_z/sample1.pdf} &
\includegraphics[width=0.15\columnwidth]{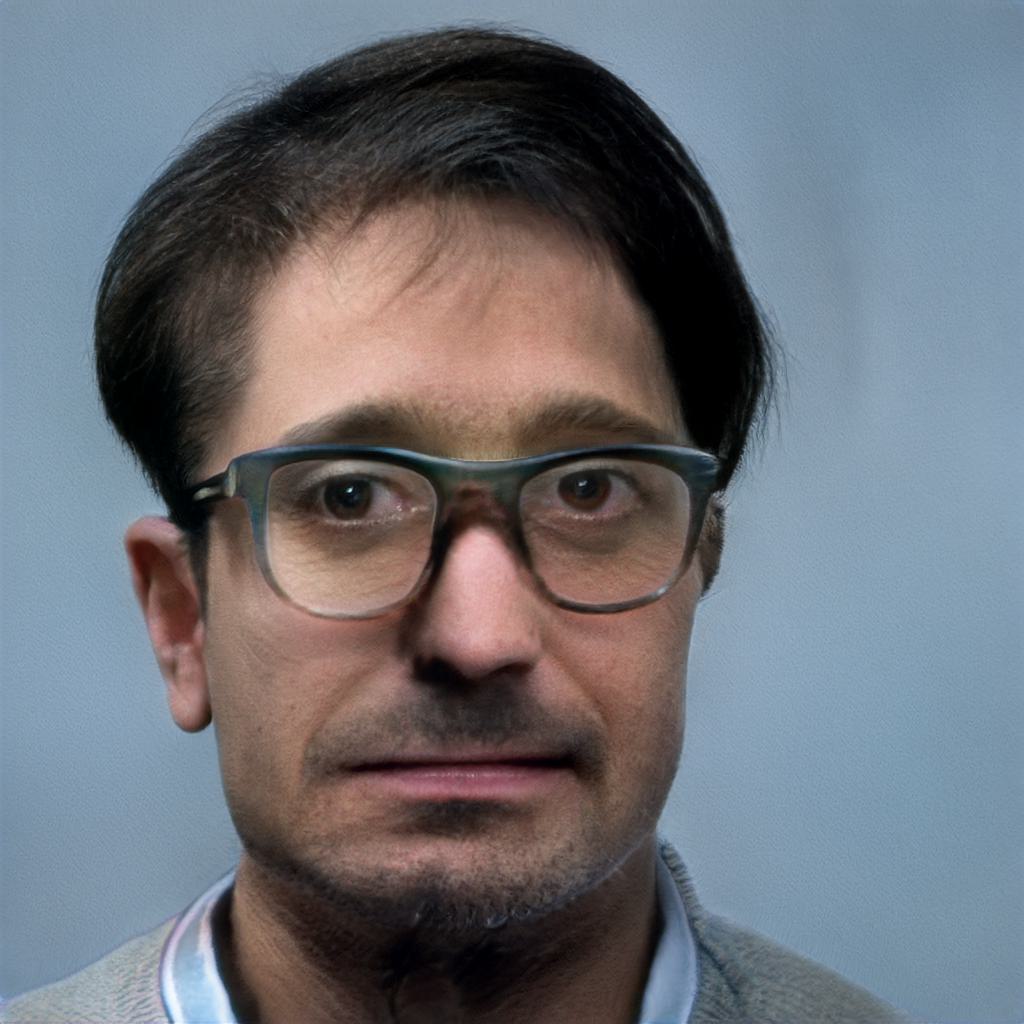} &
\includegraphics[width=0.15\columnwidth]{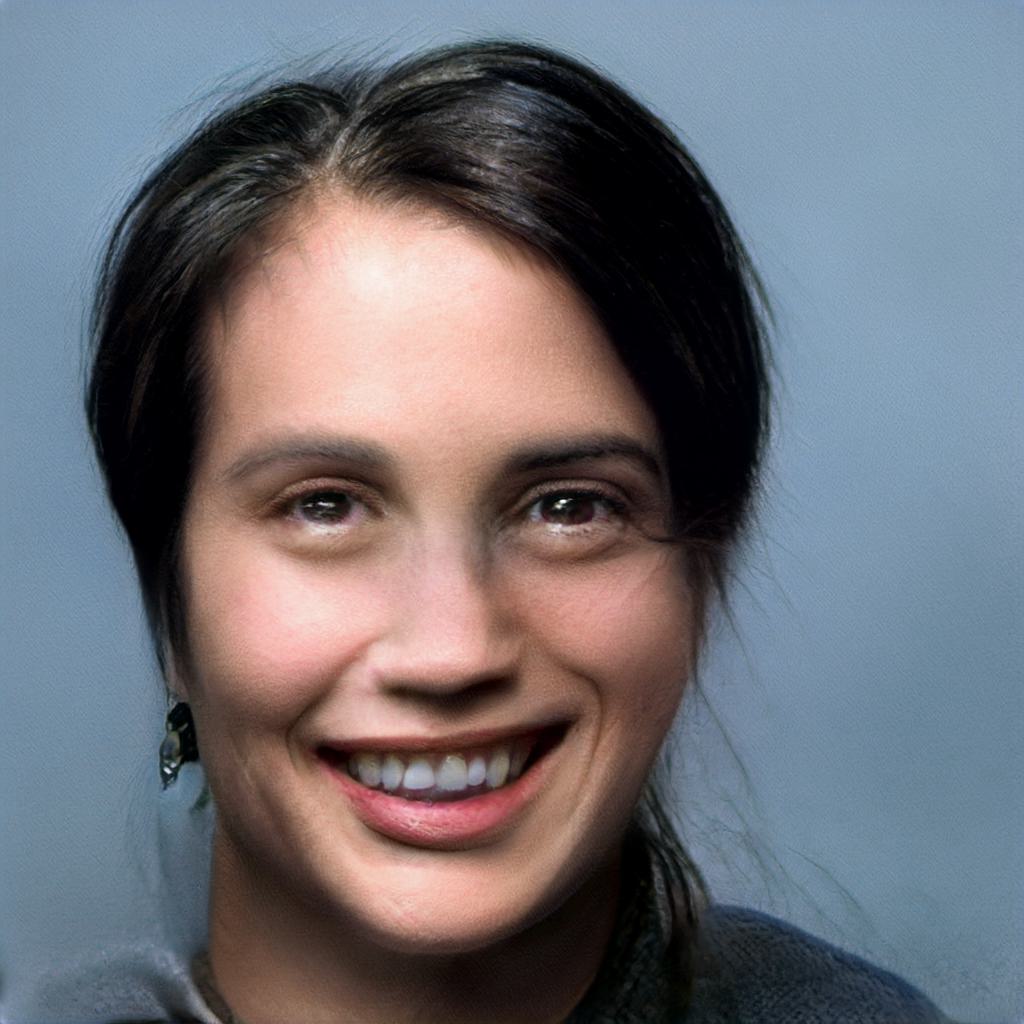} &&
\includegraphics[width=0.15\columnwidth]{rebuttal/sam_w/sample1.pdf} &
\includegraphics[width=0.15\columnwidth]{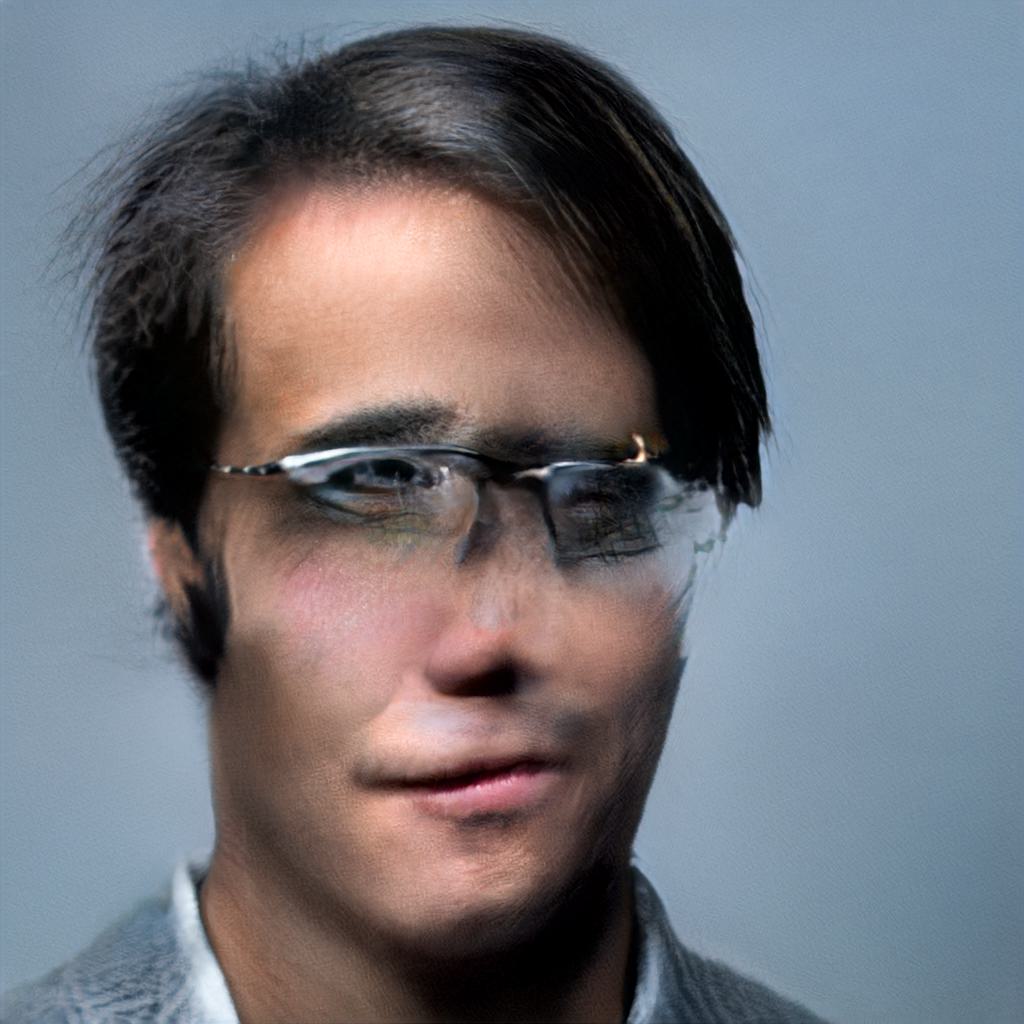} &
\includegraphics[width=0.15\columnwidth]{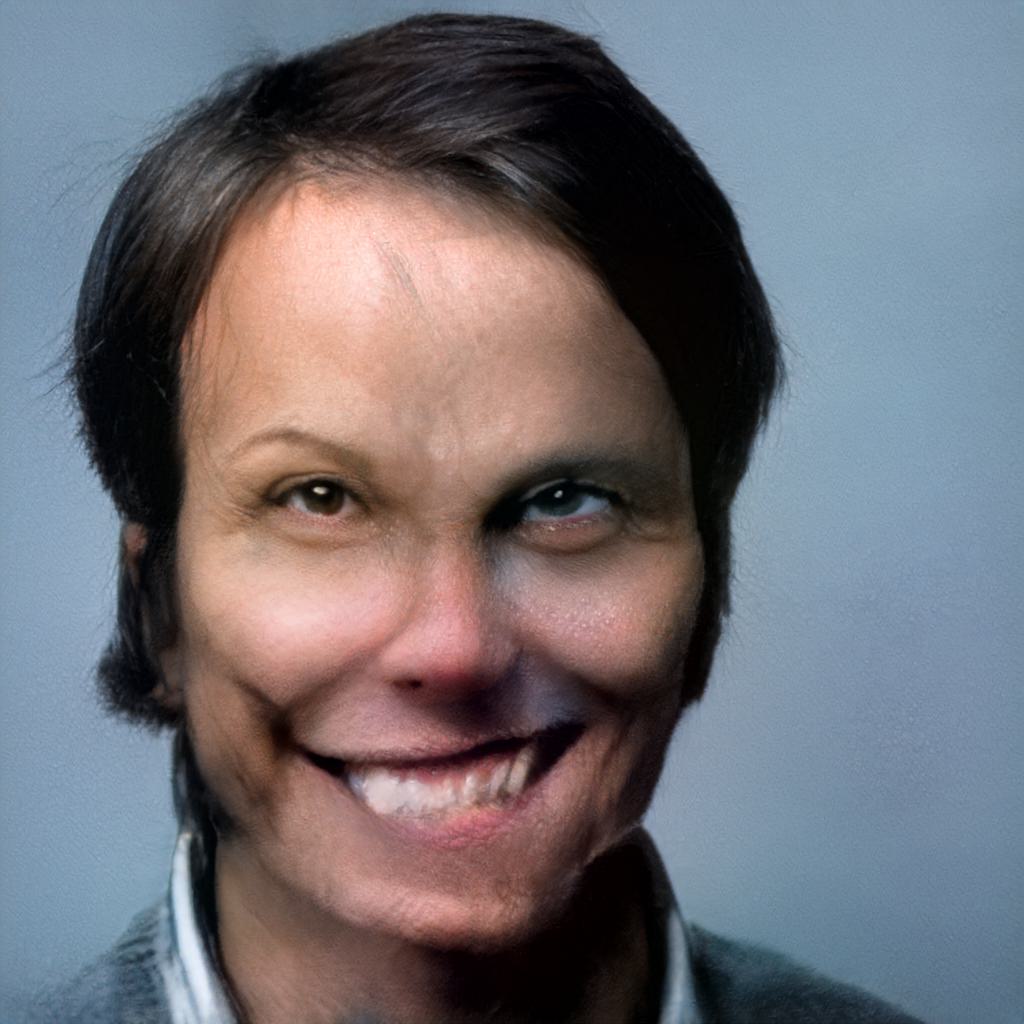} \\
\multicolumn{3}{c}{SAM~\cite{parmar2022spatially} w/ $\ZPS$} && \multicolumn{3}{c}{SAM~\cite{parmar2022spatially}} \\\\
\footnotesize inversion &\footnotesize age &\footnotesize smile &&\footnotesize inversion &\footnotesize age &\footnotesize smile \\
\includegraphics[width=0.15\columnwidth]{rebuttal/pti_z/sample5.pdf} &
\includegraphics[width=0.15\columnwidth]{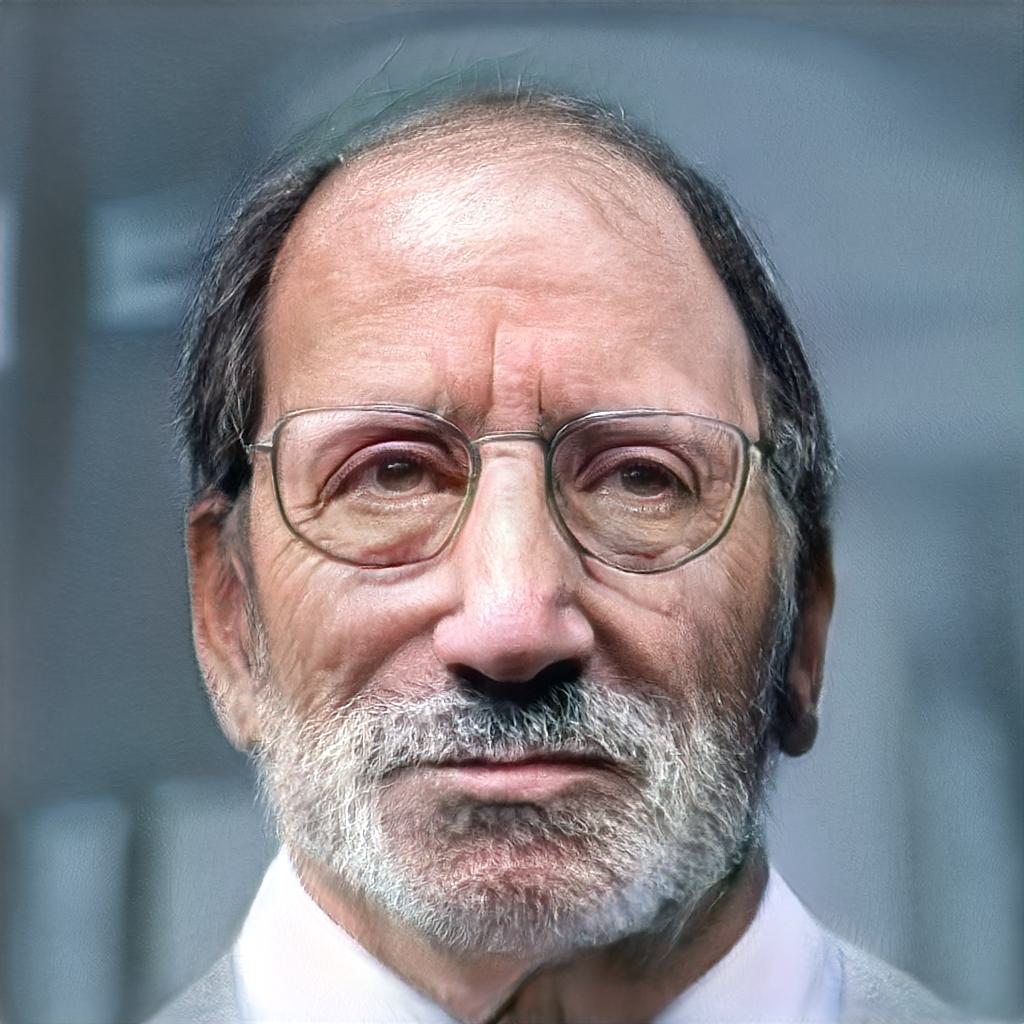} &
\includegraphics[width=0.15\columnwidth]{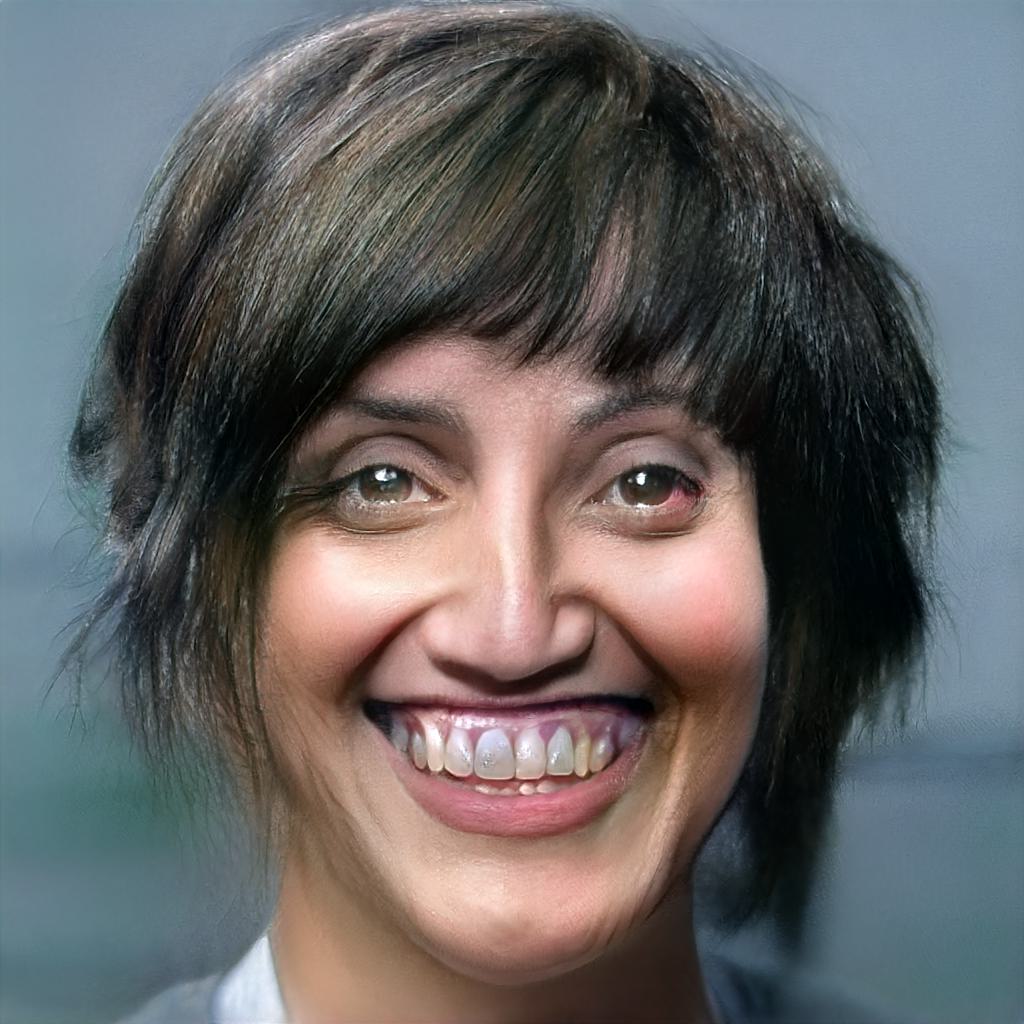} &&
\includegraphics[width=0.15\columnwidth]{rebuttal/pti_w/sample5.pdf} &
\includegraphics[width=0.15\columnwidth]{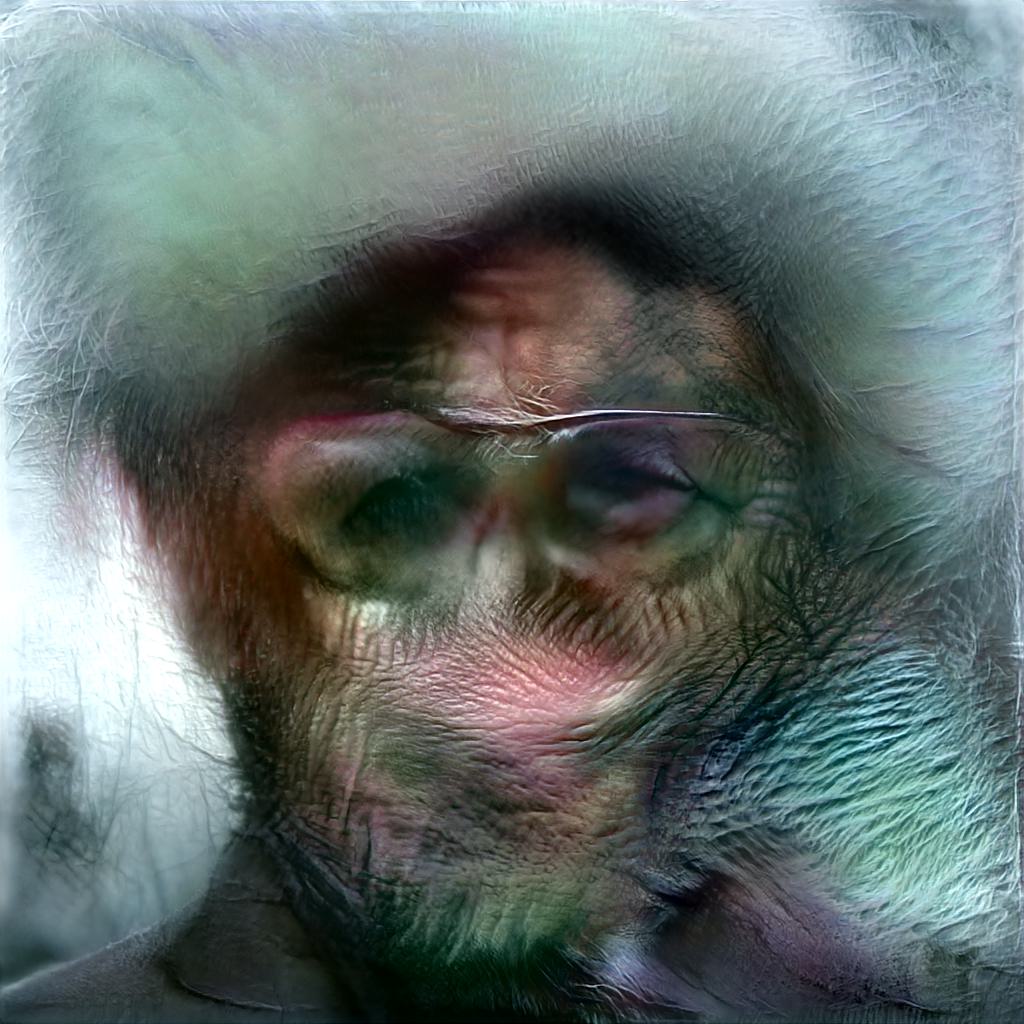} &
\includegraphics[width=0.15\columnwidth]{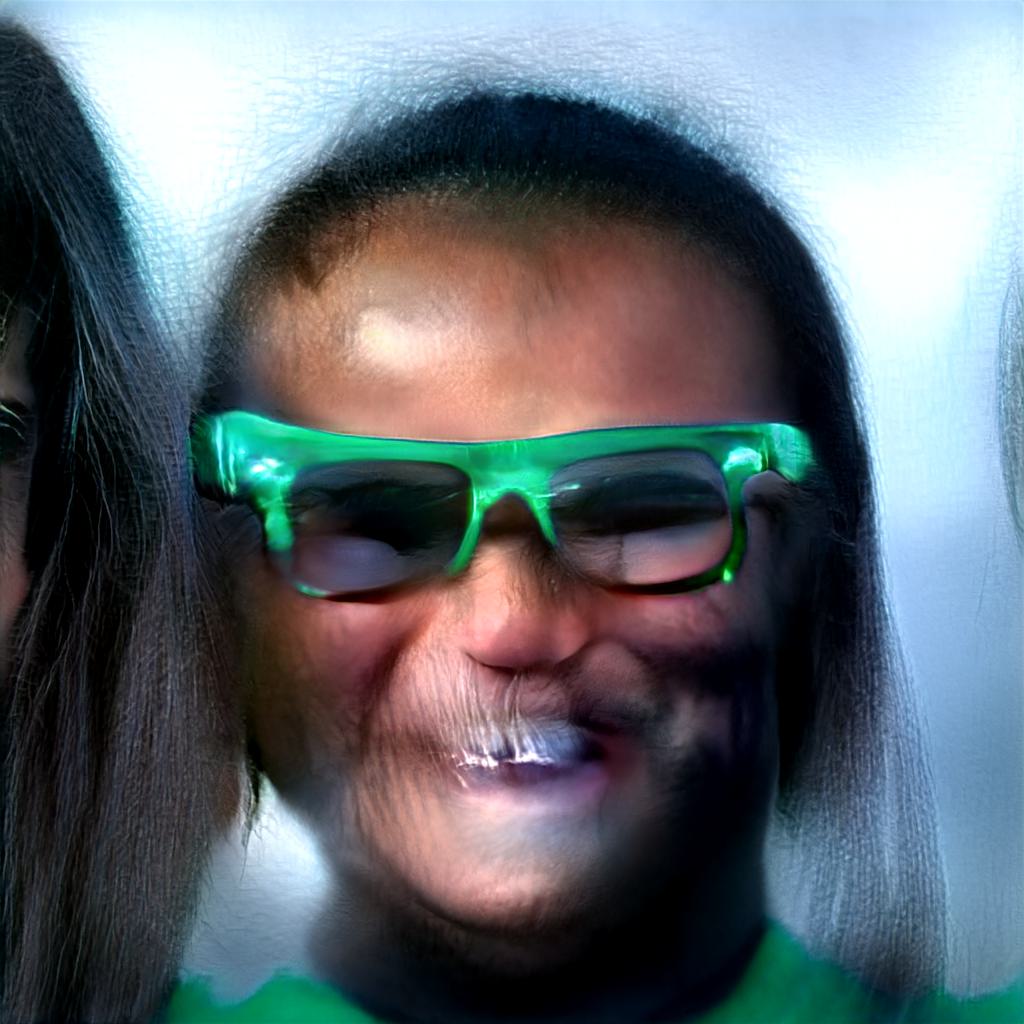} \\
\multicolumn{3}{c}{PTI~\cite{roich2021pivotal} w/ $\ZPS$} && \multicolumn{3}{c}{PTI~\cite{roich2021pivotal}}
    \end{tabular}\egroup
\vspace{-0.8em}
\caption{Editing comparison on SAM and PTI. By replacing $\WPS$ or $\WS$ to $\ZPS$ avoid harming perceptual quality of edited images.\vspace{-2em}}\label{fig:editing_sota}
\end{figure}

\noindent 
\textbf{Integration $\ZPS$ into other GAN inversion methods}.
We further demonstrate the effectiveness of our $\ZPS$ space. \Cref{fig:sota} shows reconstructed images by PTI~\cite{roich2021pivotal}, SAM~\cite{parmar2022spatially}, and $\ZPS$ version of them. We can see that the use of $\ZPS$ on PTI and SAM does not sacrifice reconstruction performance.
\Cref{fig:editing_sota} shows that integrating $\ZPS$ space into seminal GAN inversion methods relaxes editing distortions.

 \section{Conclusion}
 \label{sec:conclusion}

We revisit $\ZS$ space for GAN inversion to yield a better trade-off between reconstruction quality and editing quality. We integrate bounded latent space $\ZPS$ with the hyperspherical prior instead of $\WPS$ into the space with rich representative capacity, resulting in the presented space (\eg, $\FZS$). Our thorough experiments on PTI, SAM, $\FWS$, and $\FSS$ demonstrate that we can preserve perceptual quality of edited images while maintaining sufficient reconstruction quality on par with baseline methods by replacing unbounded space (\eg, $\WPS$) to $\ZPS$.

{\small
\bibliographystyle{ieee_fullname}
\bibliography{egbib}
}

\end{document}